\documentclass{article}
\usepackage{ijcai24}

\usepackage{times}
\usepackage{soul}
\usepackage{url}
\usepackage[hidelinks]{hyperref}
\usepackage[utf8]{inputenc}
\usepackage[small]{caption}
\usepackage{graphicx}
\usepackage{amsmath}
\usepackage{amsthm}
\usepackage{booktabs}
\usepackage{algorithm}
\usepackage{algorithmic}
\usepackage{bm}
\usepackage{dblfloatfix}
\usepackage{caption}
\usepackage{colortbl}
\usepackage{fancybox}
\usepackage{tcolorbox}
\usepackage{amssymb}

\usepackage{subcaption}
\usepackage{multicol}
\usepackage{multirow}
\usepackage[switch]{lineno}
\usepackage{pifont}
 \usepackage{dsfont}
\usepackage{sidecap}
\usepackage{footmisc}

\newcommand{\figref}[1]{Fig.~\ref{#1}}

\newcommand{\secref}[1]{Sec.~\ref{#1}}
\newcommand{\tableref}[1]{Table~\ref{#1}}

\title{\textsc{FoolSDEdit}: Deceptively Steering Your Edits Towards \\ Targeted Attribute-aware Distribution}

\renewcommand{\thefootnote}{\textasteriskcentered}
\author{
Qi Zhou$^1$\and
{Dongxia Wang$^{1,}$\thanks{corresponding authors}}\and
Tianlin Li$^2$\and
Zhihong Xu$^1$\and
Yang Liu$^2$\and
Kui Ren$^1$\and
Wenhai Wang$^1$\And
{Qing Guo$^{3,4,}$\footnotemark[1]}
\\
\affiliations
$^1$Zhejiang University, China\\
$^2$Nanyang Technological University, Singapore\\
$^3$Institute of High Performance Computing (IHPC), A*STAR, Singapore\\
$^4$Centre for Frontier AI Research (CFAR), A*STAR, Singapore\\
\emails
\{qi.zhou, dxwang, xuzhihong13, kuiren, zdzzlab\}@zju.edu.cn,
\{tianlin001, yangliu\}@ntu.edu.sg,
tsingqguo@ieee.org
}

\begin{document}
\setlength{\parindent}{1em}

\twocolumn[{
\renewcommand\twocolumn[1][]{#1}
\maketitle
\begin{center}
    \captionsetup{type=figure}
    \includegraphics[width=0.9\textwidth]{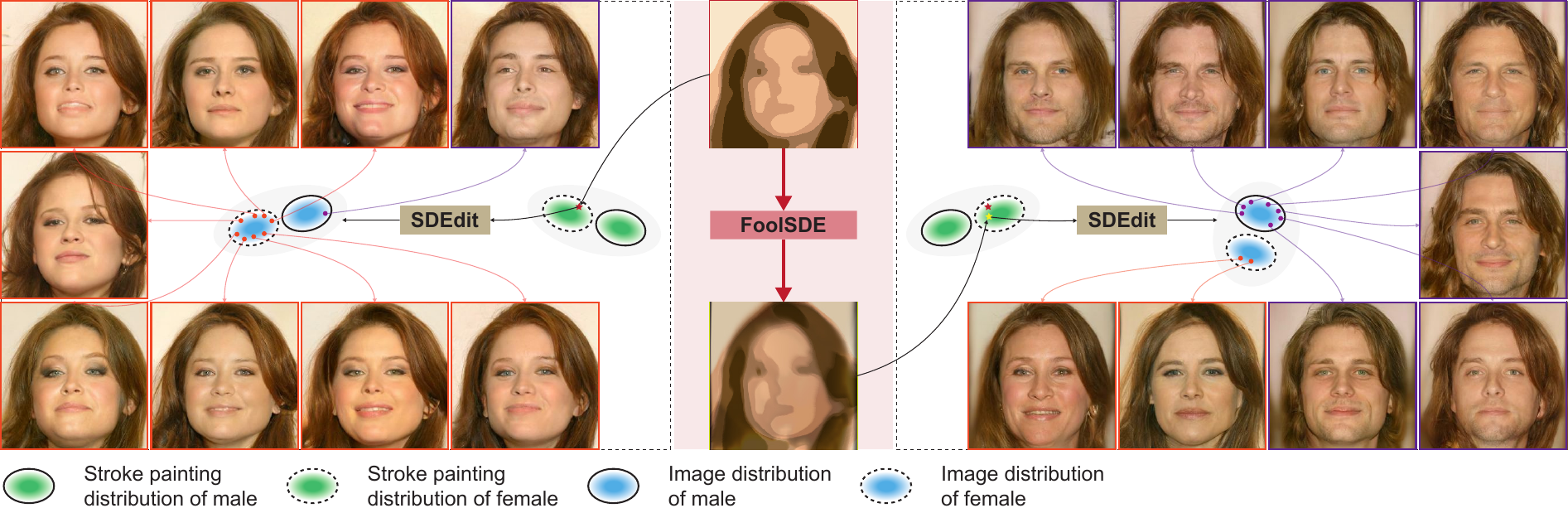}
    \captionof{figure}{Examples of our method and comparative results before and after our attack.}
    \label{fig:overview}
\end{center}
}]

\footnotetext{Corresponding authors}
\renewcommand{\thefootnote}{\arabic{footnote}}
\begin{abstract}
    Guided image synthesis methods, like SDEdit based on the diffusion model, excel at creating realistic images from user inputs such as stroke paintings.
    However, existing efforts mainly focus on image quality, often overlooking a key point: the diffusion model represents a data distribution, not individual images. This introduces a low but critical chance of generating images that contradict user intentions, raising ethical concerns. For example, a user inputting a stroke painting with female characteristics might, with some probability, get male faces from SDEdit.
    To expose this potential vulnerability, we aim to build an adversarial attack forcing SDEdit to generate a specific data distribution aligned with a specified attribute (e.g., female), without changing the input's attribute characteristics. We propose the Targeted Attribute Generative Attack (TAGA), using an attribute-aware objective function and optimizing the adversarial noise added to the input stroke painting.
    Empirical studies reveal that traditional adversarial noise struggles with TAGA, while natural perturbations like exposure and motion blur easily alter generated images' attributes.
    To execute effective attacks, we introduce \textsc{FoolSDEdit}:
    We design a joint adversarial exposure and blur attack, adding exposure and motion blur to the stroke painting and optimizing them together.
    We optimize the execution strategy of various perturbations, framing it as a network architecture search problem. We create the \textit{SuperPert}, a graph representing diverse execution strategies for different perturbations. After training, we obtain the optimized execution strategy for effective TAGA against SDEdit.
    Comprehensive experiments on two datasets show our method compelling SDEdit to generate a targeted attribute-aware data distribution, significantly outperforming baselines.
\end{abstract}

\section{Introduction}

Diffusion-based generative models can generate photo-realistic images ~\cite{ho2020denoising,song2020score,Rombach_2022_CVPR,meng2021sdedit}. 
In particular, the recently developed guided image synthesis method, \emph{i.e.}, stochastic differential editing (SDEdit) ~\cite{meng2021sdedit}, can generate realistic and faithful images according to the user inputs (\emph{e.g.}, stroke painting in \figref{fig:overview}). 
Given a guided image, SDEdit first adds noise to the image and then performs iterative denoising through the stochastic differential equation (SDE), which leads to a data distribution dependent on the guided image.  
Due to the well-balanced faithfulness and realism, SDEdit has a wide range of applications. For example, ~\cite{Gao_2023_CVPR} process images via enhanced SDEdit to adapt testing-time corruptions. 

Although achieving impressive results, existing works focus on the quality of generated individual images and neglect that SDEdit represents a data distribution based on the input-guided image.
As a data distribution, SDEdit has a probability, even though low, of generating images contradicting the user's intentions.
As illustrated in \figref{fig:overview}, when presented with a stroke painting instructing the generation of a female face, we input it into SDEdit, sampling 100 examples. Surprisingly, our findings reveal that 10\% of the generated images depict male faces, which contradicts the intended meaning specified by the stroke painting.

We argue that such a property may potentially be exploited maliciously, causing SDEdit to generate images that deviate from the user's intentions along the specified attributes.
To reveal this vulnerability, we aim to develop a novel targeted attack that adds optimized perturbations to the stroke painting without changing its raw attribute. 
When the perturbed stroke painting is input into SDEdit, the sampled examples will likely exhibit the targeted attribute.
We illustrate an example in \figref{fig:overview}. The perturbed stroke painting generated by our \textsc{FoolSDEdit} reflects the female attribute, yet the sampled images have a significant probability (approximately 80\%) of being identified as male.

To achieve the goal, we first formulate the objective as an attack task and propose the targeted attribute generative attack (TAGA). Then, we implement TAGA via the traditional adversarial additive perturbation.
TAGA is a non-trivial task, and the adversarial additive noise, even with significant strength, cannot achieve the objective.
Then, with an empirical study, we find that two natural perturbations, \emph{e.g.}, exposure variation and motion blur, could affect generated attributes easily.
To enhance the efficacy of our attacks, we introduce \textsc{FoolSDEdit}:
Initially, we devised a joint adversarial exposure and blur attack by introducing exposure and motion blur to the stroke painting, optimizing them simultaneously.
Subsequently, to fully leverage the benefits of distinct perturbations, we optimize the execution strategy of these perturbations by framing it as a network architecture search problem. Specifically, we construct the \textsc{SuperPert}, a graph capable of representing diverse execution strategies for various perturbations. We then train this graph to derive an optimized execution strategy, ensuring the effective execution of TAGA against SDEdit.
We conduct extensive experiments on two facial datasets and test the results on gender, age, and race attributes, demonstrating our method's effectiveness and advantages over baselines. 

\section{Related Work}
{\bf Guided image synthesis and editing.}
Recent advancements have applied diffusion models to edit images with impressive outcomes. SDEdit \cite{meng2021sdedit} applies and then removes noise from an image, which may include user-customized brush strokes, but it's restricted to overall image edits. DiffusionCLIP \cite{Kim_2022_CVPR} finetunes the score function in the reverse diffusion process using a CLIP loss to control the generated image attributes based on the text. DDIB \cite{su2023dual} edits images by obtaining latent source image encodings and decoding them via diffusion models to achieve image-to-image translation. Imagic \cite{kawar2023imagic} provides a text-based semantic image editing tool to operate on a single real image given a single natural language text prompt. Recently, some methods \cite{Avrahami_2022_CVPR,choi2021ilvr,nichol2021glide} are proposed to synthesize data with user-provided masks.

\noindent{\bf Bias in diffusion models.}
Diffusion models ~\cite{ho2020denoising,song2019generative,song2020score}, including Stable Diffusion ~\cite{Rombach_2022_CVPR}, tend to create biased images, often depicting stereotypes from neutral prompts. Cho et al. ~\cite{Cho2022DallEval} note a bias towards generating male figures for job-related prompts and limited skin tone diversity. Seshadri et al. ~\cite{seshadri2023bias} find that gender biases in occupational images stem from training data disparities. Bianchi et al. ~\cite{Bianchi_2023} observe that even basic prompts elicit stereotypical depictions. Luccioni et al. ~\cite{luccioni2023stable} introduce a tool to assess image sets for gender and ethnic representation, while Wang et al. ~\cite{wang2023t2iat} create a test revealing that Stable Diffusion links women more to family and men to career-related imagery. Fair Diffusion ~\cite{friedrich2023fair} is proposed to attenuate biases after the deployment of generative text-to-image models. Shen et al. ~\cite{shen2023finetuning} propose to end-to-end finetune diffusion models for fairness.

\noindent{\bf Adversarial attacks.} 
Adversarial attacks aim to craft and apply perturbations to input to deceive the target model into making incorrect predictions. Research has been conducted to explore the impact of such additive perturbations on the robustness of classifiers, examining their effects in both white-box ~\cite{goodfellow2014explaining,kurakin2018adversarial,madry2017towards,andriushchenko2020square,croce2020reliable} and black-box scenarios ~\cite{liu2016delving,cheng2019improving,shi2019curls,guo2019simple,ilyas2018black}.
Recently, researchers have also explored the use of naturally occurring degradations as forms of attack perturbations. These include environmental and processing effects like motion blur, vignetting, rain streaks, varying exposure levels, and watermarks ~\cite{gao2022can,guo2020watch,jia2020adv,tian2021ava,hou2023evading}.
To achieve adversarial attacks against diffusion models,  we investigate
the potential of natural degradation to establish an attribute distribution-aware attack method.

\section{Background and Analysis}
\label{sec:preliminary}

\begin{SCfigure*}
    \centering
    \includegraphics[scale=0.67]{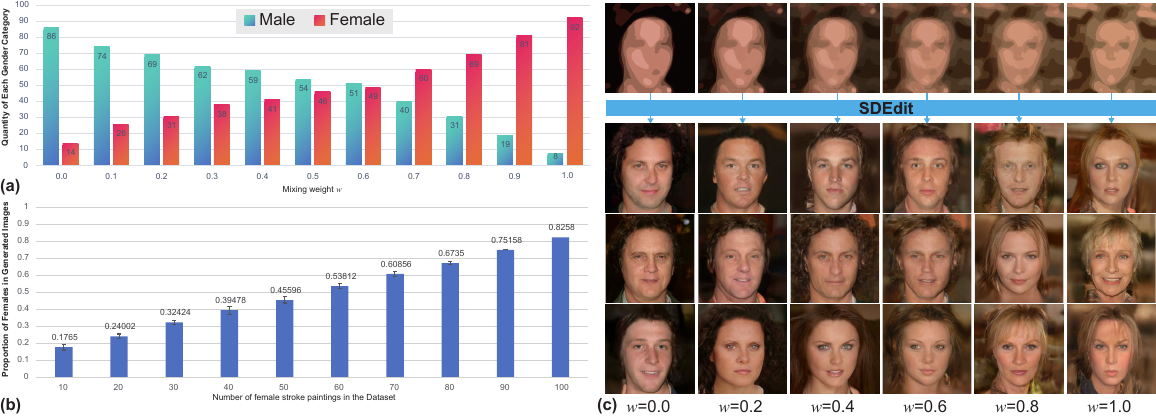}
    \caption{SDEdit analysis with different gender attribute setups. (a) We mix one female stroke painting and one male stroke painting with different weights $w$ in Eq. (\ref{eq:sdedit-distr}) and display the number of generated examples w.r.t. the gender categories. (b) We conduct dataset-level analysis by setting the different amounts of female faces. (c) We display the mixed inputs and the generated results.}
    \label{fig:analysis}
\end{SCfigure*}

\textbf{Preliminaries of SDEdit.}
SDEdit \cite{meng2021sdedit} formulates the guided image synthesis as 
a reverse stochastic differential equation (SDE) problem \cite{song2020score} that is solved from the intermediate time. Specifically, given a guided image $\mathbf{X}^\text{g}$, \emph{e.g.}, stroke painting containing coarse structure information (See \figref{fig:overview}), we can set a Gaussian distribution by regarding $\mathbf{X}^\text{g}$ as the expectation,
\begin{align} \label{eq:sdedit1}
    \mathbf{X}^\text{g}(t_0)  \sim \mathcal{N}(\mathbf{X}^\text{g};\sigma^2(t_0)\mathbf{I}).
\end{align}
Then, we regard $\mathbf{X}^\text{g}(t_0)$ as an intermediate status at the time $t_0$ in the reverse SDE process (\emph{i.e.}, $t_0\in (0,1)$) and use the standard reverse SDE \cite{song2020score} to obtain the desired data distribution by gradually removing noise from $\mathbf{X}^\text{g}(t_0)$,
\begin{align} \label{eq:sdedit2}
    \mathbf{X} \sim p_\text{data}=\text{SDEdit}(\mathbf{X}^\text{g},t_0,\theta), 
\end{align}
where $\theta$ denotes the parameter of a score model. In the following, we simplify $\text{SDEdit}(\mathbf{X}^\text{g},t_0,\theta)$ as $\text{SDEdit}(\mathbf{X}^\text{g})$ since we set the same $t_0$ and $\theta$ for all inputs.
Previous works mainly focus on how to generate realistic and faithful individual examples but neglect that SDEdit is to generate a data distribution. 
The individual examples are sampled from this distribution.
As a distribution, the sampled examples have a probability of contradicting the input's intentions. 

\noindent\textbf{SDEdit can generate unexpected examples along a specific attribute.}
We designed an experiment to validate the above fact and take the gender attribute as an example. 
Specifically, given two strokes of two human faces, one is female and another is male, which are generated from a real female and male face images and denoted as $\mathbf{X}^\text{g}_\text{f}$ and $\mathbf{X}^\text{g}_\text{m}$, respectively. 
Then, we mix the two stroke paintings with different mixing weights and get augmented paintings
\begin{align}\label{eq:sdedit-distr}
    \mathbf{X}^\text{g}_\text{mix}(w) = w\mathbf{X}^\text{g}_\text{f} + (1-w)\mathbf{X}^\text{g}_\text{m},
\end{align}
where $w\in [0,1]$. Then, we normally sample ten weights within the range and get ten strokes that transfer from male characteristics to female characteristics gradually. 
We feed the ten stroke paintings into Eq. (\ref{eq:sdedit2}) and get ten data distributions denoted as $\{p_\text{data}(w)\}$, respectively. 
Note that such a mixing strategy has been used as a data augmentation in the fairness enhancement \cite{mroueh2021fair}, demonstrating its reasonability.
After that, for each distribution, we sample 100 examples randomly and feed these examples to a CLIP-based gender classification.
Finally, we can count the numbers of females and males. 
We show the results in \figref{fig:analysis}(a) and (c) and observe that: 
\ding{182} The ratio of sampled female faces gradually increases as the mixing weight $w$ becomes larger. 
This observation demonstrates that the generated distributions of SDEdit follow the inputs' intentions according to the gender attribute.   
\ding{183} There is a probability of generating examples contradicting the input's intentions.
When we use the raw stroke painting (\emph{e.g.}, $w=0.0$) that aims to generate male faces, we can still synthesize 14\% female faces sampled from the data distribution generated by $\text{SDEdit}(\mathbf{X}^\text{g}_\text{mix}(w=0.0))$. 

Beyond the above individual examples, we also conduct a dataset-level analysis. 
Specifically, we build a set of stroke paintings denoted as $\mathcal{S}(N)$ that contains $|\mathcal{S}|$ stroke paintings. The $N$ of these stroke paintings are female characteristics, and the rest of the stroke paintings are male characteristics. 
We can adjust the $N$ and get different stroke painting datasets denoted as $\{\mathcal{S}(N)\}$.
For each stroke painting in $\mathcal{S}(N)$, we feed it into Eq. (\ref{eq:sdedit2}), generating 100 examples. We then calculate the ratio of generated female faces for each painting and obtain the average ratio across all stroke paintings.
Finally, we depict the findings in a histogram in \figref{fig:analysis}, revealing a noteworthy trend: as the number of female stroke paintings increases, the average ratio of generated female faces also rises. 
Surprisingly, even when all stroke paintings are of females, the method still yields an average of 17.42\% male faces (as indicated by the last bar in \figref{fig:analysis}(b)).

\textit{
Based on the preceding analysis, SDEdit demonstrates a high probability of generating examples that align with the desired attributes from the input stroke painting. However, it also can generate examples that contradict the intended characteristics of the stroke painting.
}
In this study, our objective is to explore whether this inherent property can be accentuated, leading SDEdit to generate examples that deliberately diverge from the intended attributes of the stroke painting. Such a scenario could give rise to security concerns.

\section{Targeted Attribute Generative Attack}
\label{sec:targetedattack}



\subsection{Problem Formulation} 
\label{subsec:problem_formulation}

During the guided image synthesis, users provide a guided image $\mathbf{X}^\text{g}\in \mathds{R}^{H\times W}$ with a specified attribute denoted as $a$ and express their purposes to synthesize examples along $a$  (\emph{e.g.}, gender, age, race, etc.). For example, in \figref{fig:overview}, the stroke painting means the users may want a female's face and $a =\text{`female'}$.
The SDEdit is to generate a data distribution $p_\text{data}$ according to $\mathbf{X}^\text{g}$ and Eq. (\ref{eq:sdedit2}). The generated examples are desired to have the same attribute with $a$.

In this work, our objective is to embed some perturbations in the $\mathbf{X}^\text{g}$ and get $\tilde{\mathbf{X}}^\text{g}=f_\beta(\mathbf{X}^\text{g})$ without changing the attribute $a$, where $f_\beta(\cdot)$ is the function to add perturbations to the guided image with the parameters as $\beta$.
Then, we feed $\tilde{\mathbf{X}}^\text{g}$ into Eq. (\ref{eq:sdedit2}) to generate a new data distribution $\tilde{p}^\text{data}$ that can synthesize examples with a high probability to have the targeted attribute $\hat{a}$ instead of the attribute $a$ in $\tilde{\mathbf{X}}^\text{g}$.
For example, in \figref{fig:overview}, the perturbed stroke painting makes SDEdit tend to generate eight male faces instead of the female face.
We can formulate the attack objective as follows
\begin{align} \label{eq:problem}
    & \arg\min_{\beta} \mathcal{J}(\text{SDEdit}(f_\beta(\mathbf{X}^\text{g})),\hat{a}), \nonumber\\
    & \text{subject to}~|\beta|_p \leq \epsilon, f_\text{att}(f_\beta(\mathbf{X}^\text{g}))) = f_\text{att}(\mathbf{X}^\text{g}),
\end{align}
where $\mathcal{J}(\cdot)$ is the objective function to measure whether $\text{SDEdit}(f_\beta(\mathbf{X}^\text{g}))$ fits the targeted attribute $\hat{a}$, and $f_\text{att}(\cdot)$ is to extract the attribute of input. 
The perturbation $\beta$ should be within the $L_p$ bound (\emph{i.e.}, $\epsilon$), similar to existing attacks.
The constraint $f_\text{att}(f_\beta(\mathbf{X}^\text{g})) = f_\text{att}(\mathbf{X}^\text{g})$ is to make sure the added perturbation does not change the attribute of $\mathbf{X}^\text{g}$. 

\begin{figure}[t]
    \centering
    \includegraphics[width=\linewidth]{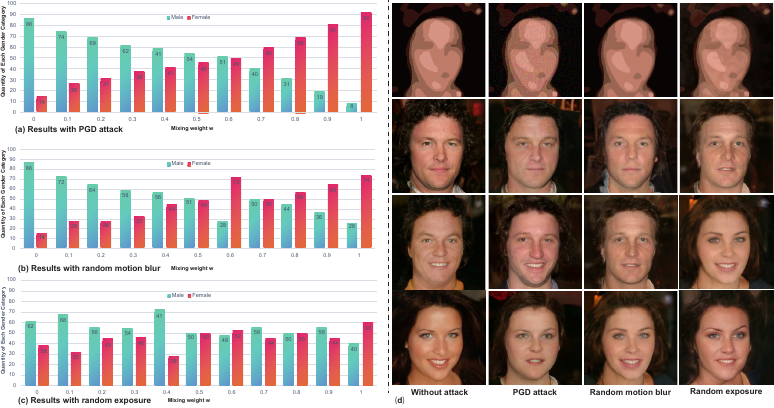}
    \caption{
    Adding adversarial noise from PGD, random motion blur, and random exposure to the mixing stroke paintings in \figref{fig:analysis} (a). 
    }
    \label{fig:study}
\end{figure}

\subsection{Naive Implementation} 
\label{subsec:naiveimpl}

We can implement Eq. (\ref{eq:problem}) via the additive perturbation attack and optimize the perturbation following the projected gradient descent (PGD). To this end, we have the following setups.

\noindent\textbf{Setup of $f_\beta(\cdot)$.} We use the additive perturbation for $f_\beta(\cdot)$ and have $f_\beta(\mathbf{X}^\text{g})=\tilde{\mathbf{X}}^\text{g}=\mathbf{X}^\text{g}+\beta$ where $\beta$ has the same size with $\mathbf{X}^\text{g}$.

\noindent\textbf{Setup of $\mathcal{J}(\cdot)$.} We define the objective function as follows
\begin{align}\label{eq:taga_obj}
    \mathcal{J}(\text{SDEdit}(f_\beta(\mathbf{X}^\text{g}))) = \sum_{\mathbf{X}\in p_\text{data}} \mathcal{J}_\text{cross}(f_\text{att}(\mathbf{X}),\hat{a}),
\end{align}
where $p_\text{data}=\text{SDEdit}(f_\beta(\mathbf{X}^\text{g}))$, $\mathcal{J}_\text{cross}(\cdot)$ is the cross-entropy loss function, $\hat{a}$ is the targeted attribute we want the generated images should have.

\noindent\textbf{Setup of $f_{\text{att}}(\cdot)$.} We adopt a CLIP-based gender classifier for its superior performance in zero-shot face attribute classification tasks \cite{radford2021learning}. Specifically, followed by the original work, we set the text prompts as ``\textit{A photo of a \{\} person}" for the CLIP.

\noindent\textbf{Other setups.} We use the $L_\infty$ norm to restrict the noise strength and set $\epsilon=8/255$.

\subsection{Empirical Study and Motivation} 
\label{subsec:analysis}

In this section, we conduct some preliminary studies and find that it is difficult to achieve the goal defined in \secref{subsec:problem_formulation} through the naive implementation in \secref{subsec:naiveimpl}. 

Specifically, we can get adversarial additive noise via PGD attack through the setup in \secref{subsec:naiveimpl}, and we denote it as PGD. Meanwhile, we propose to add random motion blur and exposure effect to the stroke painting to study the influence of the natural degradations. 
Specifically, we add the three types of perturbations to the stroke paintings calculated from Eq. (\ref{eq:sdedit-distr}) and get three histograms similar to \figref{fig:analysis} (a).
As shown in \figref{fig:study}, PGD attack does not change the main shape the histogram when we compare it to \figref{fig:analysis} (a). 
In contrast, we see that both random motion blur and exposure could affect the ratios of female and male effectively.
All of the changed stroke paintings lead to natural and realistic generations.
The results inspire us to conduct adversarial blur and exposure attacks to achieve TAGA based on the following reasons: 
\ding{182} Compared with the additive noise, motion blur and exposure are common degradations that usually appear in the real world and adding them to stroke paintings could represent the cases that may happen in the real world. 
\ding{183} Both perturbations affect the ratios of female and male significantly without changing the attribute in the stroke paintings.

\section{\textsc{FoolSDEdit}}
\label{sec:foolsdedit}

In this section, we propose the \textsc{FoolSDEdit} to realize the targeted attribute generative attack in \secref{sec:targetedattack}. Inspired by the analysis in \secref{subsec:analysis}, we propose to add the adversarially optimized exposure and blur to stroke painting in \secref{subsec:jointTAGA}. 
To allow an effective attack, we extend the combination to a deep architecture denoted as \textsc{SuperPert} in \secref{subsec:superpert} that allows added multi-layer adversarial exposure and blur in a parallel way. 
Finally, we detail how to optimize the architecture and perturbation parameters jointly. 

\subsection{Joint Adv. Exposure and Blur for TAGA}
\label{subsec:jointTAGA}

To allow adversarial exposure for TAGA, we set the exposure-based perturbation function $f_{\beta_\text{e}}(\mathbf{X}^\text{g})$ as
\begin{align}\label{eq:jointTAGA-exp}
    f_{\beta_\text{e}}(\mathbf{X}^\text{g}) = \mathbf{X}^\text{g}\odot \beta_\text{e} = \log^{-1}(\mathbf{X}^{\text{g}'}+ \beta_\text{e}'),
\end{align}
where $\beta_\text{e}\in \mathds{R}^{H\times W}$ is the pixel-wise exposure matrix, and $\mathbf{X}^{\text{g}'}$ and $\beta_\text{e}'$ are their logarithmic versions.
To allow adversarial blur for TAGA, we set 
\begin{align}\label{eq:jointTAGA-blur}
    f_{\beta_\text{b}}(\mathbf{X}^\text{g}) = \beta_\text{b} \circledast \mathbf{X}^\text{g},
\end{align}
where `$\circledast $' denotes the pixel-wise filtering, that is, each pixel of $\mathbf{X}^\text{g}$ is filtered by the kernel stored at the corresponding location of $\beta_\text{b} \in \mathds{R}^{H\times W \times K^2}$.
With $f_{\beta_\text{b}}(\mathbf{X}^\text{g})$ and $f_{\beta_\text{e}}(\mathbf{X}^\text{g})$, we can define the $f_{\beta}(\mathbf{X}^\text{g})$ in Eq. (\ref{eq:problem}) in several ways. 
E.g., we can execute them in a serial way, like  $f_\beta(\mathbf{X}^\text{g}) = f_{\beta_\text{e}} (f_{\beta_1}(\mathbf{X}^\text{g}))$ or a parallel way, like  
$f_\beta(\mathbf{X}^\text{g}) = w_1f_{\beta_1}(\mathbf{X}^\text{g}) + w_2f_{\beta_2}(\mathbf{X}^\text{g})$. 
Moreover, we can also execute the same perturbation for multiple times.
Then, how to find a good execution way is a key question.
Once we confirm the execution way to conduct the two kinds of perturbations, we can use the \secref{subsec:naiveimpl} to realize the attack. 
How to optimize the execution way and perturbation parameters jointly should also be studied.

\subsection{\textbf{\textsc{SuperPert}}}
\label{subsec:superpert}

\begin{figure}[t]
    \centering
    \includegraphics[width=\linewidth]{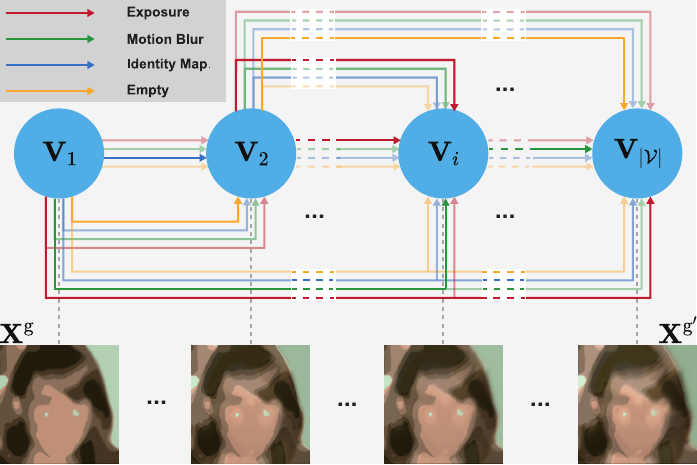}
    \caption{
    Pipeline of \textsc{SuperPert}. $\mathbf{X}^\text{g}$ is the input stroke painting and $\mathbf{X}^{\text{g}'}$ is the adversarial perturbed stroke painting.
    }
    \label{fig:archsearch}
\end{figure}

To find a good execution way for $f_\beta(\cdot)$, we formulate it as an architecture searching problem similar to \cite{liu2018darts} and propose the \textsc{SuperPert}.
Specifically, we set a graph $\mathcal{G}=(\mathcal{V},\mathcal{E})$ to represent the whole architecture. 
The node $\mathbf{V}_i\in \mathcal{V}$ represents an intermediate image generated in the process, and the edge $\mathbf{E}_{i\rightarrow j}\in \mathcal{E}$ denotes an operation that maps the node $\mathbf{V}_i$ to $\mathbf{V}_j$.
We set an operation set $\mathcal{O}$ including four operations, \emph{i.e.}, $\mathcal{O}=\{\text{O}^1,\text{O}^2, \text{O}^3, \text{O}^4\}$, corresponding to exposure, motion blur, identity mapping, and empty operation outputting zero image, respectively.
For each edge, we set a weight $\alpha_{i\rightarrow j}^k$ to represent the
contribution of the $k$th operation in $\mathcal{O}$ to a well-performed architecture.
We can represent the edge $\mathbf{E}_{i\rightarrow j}$ as follows
\begin{align}\label{eq:superpert-edge}
    \mathbf{V}_j = \sum_{k=1}^4 \alpha_{i\rightarrow j}^k \text{O}^k(\mathbf{V}_i).
\end{align}
We use the graph to represent the whole architecture to process the input $\mathbf{X}^\text{g}$ and display the graph in \figref{fig:archsearch}.
All weights form a set, \emph{i.e.}, $\mathcal{A}=\{\alpha_{i\rightarrow j}^k\}$, which represent the architecture searching space.
All perturbation-related parameters form a set $\mathcal{B}=\{\beta_{i\rightarrow j}^k\}$ where $\beta_{i\rightarrow j}^k$ denotes the perturbation parameter for the $k$th operation at the edge $\mathbf{E}_{i\rightarrow j}$.
We aim to optimize $\mathcal{A}$ and get a finalized architecture for attacking.

\begin{table*}[h]
\centering
\caption{The overview results of different attack methods on gender attribute.}
\label{tab:table_gender}
\vspace{-10pt}
\setlength{\tabcolsep}{5.5pt}
\resizebox{1\linewidth}{!}{
\renewcommand\arraystretch{1}
\begin{tabular}{ccl|cccc|cccc|cccc|cccc}
\hline 
\multicolumn{1}{l}{} &
  \multicolumn{1}{l}{} &
   &
  \multicolumn{4}{c|}{SuperPert} &
  \multicolumn{4}{c|}{Auto-PGD} &
  \multicolumn{4}{c|}{ODI-PGD} &
  \multicolumn{4}{c}{Jitter} \\
\multicolumn{1}{l}{} &
  \multicolumn{1}{l}{} &
   &
  AADS$\textcolor{red}{\uparrow}$ &
  ACADS$\textcolor{blue}{\downarrow}$ &
  AoCC$\textcolor{red}{\uparrow}$ &
  $d_\text{NIQE}\textcolor{blue}{\downarrow}$ &
  AADS$\textcolor{red}{\uparrow}$ &
  ACADS$\textcolor{blue}{\downarrow}$ &
  AoCC$\textcolor{red}{\uparrow}$ &
  $d_\text{NIQE}\textcolor{blue}{\downarrow}$ &
  AADS$\textcolor{red}{\uparrow}$ &
  ACADS$\textcolor{blue}{\downarrow}$ &
  AoCC$\textcolor{red}{\uparrow}$ &
  $d_\text{NIQE}\textcolor{blue}{\downarrow}$ &
  AADS$\textcolor{red}{\uparrow}$ &
  ACADS$\textcolor{blue}{\downarrow}$ &
  AoCC$\textcolor{red}{\uparrow}$ &
  $d_\text{NIQE}\textcolor{blue}{\downarrow}$ \\ \hline \hline 
\multicolumn{1}{c|}{} &
  \multicolumn{1}{c|}{} &
  hat &
  \textbf{$_{(97.70\%-)}$\hspace{4pt}5.10\%} &
  \hspace{8pt}3.21\% &
  100\% &
  $_{(5.63)}$-0.53 &
  \hspace{4pt}0.90\% &
  \hspace{4pt}15.62\% &
  \hspace{4pt}70\% &
  -0.23 &
  -0.10\% &
  \hspace{4pt}4.34\% &
  \hspace{4pt}90\% &
  -0.21 &
  \hspace{4pt}0.70\% &
  \hspace{4pt}17.83\% &
  \hspace{4pt}50\% &
  -0.22 \\
\multicolumn{1}{r|}{} &
  \multicolumn{1}{c|}{} &
  no hat &
  \textbf{$_{(81.70\%-)}$18.10\%} &
  \hspace{8pt}0.65\% &
  100\% &
  $_{(5.83)}$-0.59 &
  \hspace{4pt}5.10\% &
  \hspace{4pt}28.91\% &
  \hspace{4pt}40\% &
  -0.03 &
  \hspace{4pt}3.10\% &
  15.37\% &
  \hspace{4pt}70\% &
  -0.02 &
  \hspace{4pt}5.10\% &
  \hspace{4pt}25.95\% &
  \hspace{4pt}60\% &
  -0.02 \\
\multicolumn{1}{c|}{} &
  \multicolumn{1}{c|}{} &
  glasses &
  \textbf{$_{(94.30\%-)}$10.40\%} &
  \hspace{4pt}-6.02\% &
  100\% &
  $_{(5.31)}$-0.15 &
  \hspace{4pt}6.30\% &
  \hspace{4pt}14.23\% &
  \hspace{4pt}60\% &
  -0.10 &
  \hspace{4pt}4.50\% &
  10.68\% &
  \hspace{4pt}90\% &
  -0.10 &
  \hspace{4pt}6.30\% &
  \hspace{4pt}10.18\% &
  \hspace{4pt}80\% &
  -0.10 \\
\multicolumn{1}{c|}{} &
  \multicolumn{1}{c|}{} &
  no glasses &
  \textbf{$_{(80.30\%-)}$\hspace{4pt}7.70\%} &
  \hspace{8pt}0.28\% &
  100\% &
  $_{(5.36)}$-0.11 &
  \hspace{4pt}5.70\% &
  \hspace{4pt}23.86\% &
  \hspace{4pt}60\% &
  -0.01 &
  \hspace{4pt}2.20\% &
  13.93\% &
  \hspace{4pt}90\% &
  -0.01 &
  \hspace{4pt}5.40\% &
  \hspace{4pt}24.72\% &
  \hspace{4pt}60\% &
  -0.00 \\
\multicolumn{1}{c|}{} &
  \multicolumn{1}{c|}{} &
  oval &
  \textbf{$_{(90.50\%-)}$19.50\%} &
  \hspace{4pt}-2.35\% &
  100\% &
  $_{(5.73)}$+0.05 &
  \hspace{4pt}7.00\% &
  \hspace{8pt}3.33\% &
  \hspace{4pt}90\% &
  +0.03 &
  \hspace{4pt}5.50\% &
  \hspace{4pt}6.79\% &
  \hspace{4pt}90\% &
  +0.02 &
  -7.30\% &
  \hspace{4pt}-9.23\% &
  \hspace{4pt}80\% &
  -0.34 \\
\multicolumn{1}{c|}{} &
  \multicolumn{1}{c|}{\multirow{-6}{*}{M}} &
  non oval &
  \textbf{$_{(78.70\%-)}$\hspace{4pt}8.80\%} &
  \hspace{8pt}4.54\% &
  100\% &
  $_{(5.16)}$-0.09 &
  \hspace{4pt}7.00\% &
  \hspace{4pt}27.37\% &
  \hspace{4pt}60\% &
  -0.00 &
  \hspace{4pt}3.00\% &
  16.75\% &
  \hspace{4pt}70\% &
  -0.01 &
  \hspace{4pt}6.30\% &
  \hspace{4pt}26.59\% &
  \hspace{4pt}30\% &
  +0.55 \\ \cline{2-19} 
\multicolumn{1}{c|}{} &
  \multicolumn{2}{c|}{\cellcolor[HTML]{FFE599}Male Avg} &
  \cellcolor[HTML]{FFE599}\textbf{$_{(87.23\%-)}$11.60\%} &
  \cellcolor[HTML]{FFE599}\hspace{8pt}0.05\% &
  \cellcolor[HTML]{FFE599}100\% &
  \cellcolor[HTML]{FFE599}$_{(5.50)}$-0.12 &
  \cellcolor[HTML]{FFE599}\hspace{4pt}5.33\% &
  \cellcolor[HTML]{FFE599}\hspace{4pt}18.89\% &
  \cellcolor[HTML]{FFE599}\hspace{4pt}63\% &
  \cellcolor[HTML]{FFE599}-0.06 &
  \cellcolor[HTML]{FFE599}\hspace{4pt}3.03\% &
  \cellcolor[HTML]{FFE599}11.31\% &
  \cellcolor[HTML]{FFE599}\hspace{4pt}83\% &
  \cellcolor[HTML]{FFE599}-0.06 &
  \cellcolor[HTML]{FFE599}\hspace{4pt}2.75\% &
  \cellcolor[HTML]{FFE599}\hspace{4pt}16.01\% &
  \cellcolor[HTML]{FFE599}\hspace{4pt}60\% &
  \cellcolor[HTML]{FFE599}-0.02 \\ \cline{2-19} 
\multicolumn{1}{c|}{} &
  \multicolumn{1}{c|}{} &
  hat &
  \textbf{$_{(91.40\%-)}$13.90\%} &
  \hspace{8pt}8.29\% &
  100\% &
  $_{(5.70)}$-0.06 &
  \hspace{4pt}4.20\% &
  -20.61\% &
  \hspace{4pt}90\% &
  -0.11 &
  \hspace{4pt}3.40\% &
  20.32\% &
  \hspace{4pt}90\% &
  -0.09 &
  \hspace{4pt}4.10\% &
  -20.36\% &
  \hspace{4pt}90\% &
  -0.11 \\
\multicolumn{1}{c|}{} &
  \multicolumn{1}{c|}{} &
  no hat &
  \textbf{$_{(72.40\%-)}$13.70\%} &
  \hspace{8pt}0.64\% &
  \hspace{4pt}90\% &
  $_{(5.58)}$-0.33 &
  10.80\% &
  \hspace{4pt}11.80\% &
  \hspace{4pt}90\% &
  -0.14 &
  \hspace{4pt}7.60\% &
  \hspace{4pt}4.32\% &
  \hspace{4pt}80\% &
  -0.11 &
  10.70\% &
  -12.18\% &
  \hspace{4pt}90\% &
  -0.14 \\
\multicolumn{1}{c|}{} &
  \multicolumn{1}{c|}{} &
  glasses &
  \textbf{$_{(91.90\%-)}$22.10\%} &
  \hspace{8pt}5.89\% &
  100\% &
  $_{(6.10)}$-0.16 &
  \hspace{4pt}4.40\% &
  \hspace{4pt}-7.06\% &
  100\% &
  -0.05 &
  \hspace{4pt}5.20\% &
  \hspace{4pt}3.31\% &
  100\% &
  -0.08 &
  \hspace{4pt}5.70\% &
  \hspace{8pt}7.00\% &
  100\% &
  -0.12 \\
\multicolumn{1}{c|}{} &
  \multicolumn{1}{c|}{} &
  no glasses &
  \textbf{$_{(84.50\%-)}$30.40\%} &
  \hspace{8pt}6.94\% &
  \hspace{4pt}80\% &
  $_{(5.73)}$-0.23 &
  15.00\% &
  \hspace{4pt}11.03\% &
  \hspace{4pt}60\% &
  -0.19 &
  13.20\% &
  \hspace{4pt}0.36\% &
  \hspace{4pt}80\% &
  -0.16 &
  15.00\% &
  \hspace{4pt}10.47\% &
  \hspace{4pt}90\% &
  -0.19 \\
\multicolumn{1}{c|}{} &
  \multicolumn{1}{c|}{} &
  oval &
  \textbf{$_{(93.30\%-)}$\hspace{4pt}5.80\%} &
  \hspace{4pt}-4.63\% &
  100\% &
  $_{(6.11)}$-0.21 &
  \hspace{4pt}2.80\% &
  \hspace{4pt}-2.85\% &
  100\% &
  +0.05 &
  \hspace{4pt}1.10\% &
  -9.31\% &
  100\% &
  +0.07 &
  \hspace{4pt}4.20\% &
  \hspace{8pt}7.41\% &
  \hspace{4pt}90\% &
  -0.54 \\
\multicolumn{1}{c|}{} &
  \multicolumn{1}{c|}{\multirow{-6}{*}{F}} &
  non oval &
  \textbf{$_{(68.00\%-)}$15.20\%} &
  \hspace{4pt}-1.34\% &
  \hspace{4pt}80\% &
  $_{(5.76)}$-0.38 &
  \hspace{4pt}9.10\% &
  -15.05\% &
  100\% &
  -0.23 &
  \hspace{4pt}8.00\% &
  -1.87\% &
  \hspace{4pt}90\% &
  -0.15 &
  \hspace{4pt}4.30\% &
  -33.69\% &
  \hspace{4pt}90\% &
  -0.32 \\ \cline{2-19} 
\multicolumn{1}{c|}{\multirow{-14}{*}{CelebA}} &
  \multicolumn{2}{c|}{\cellcolor[HTML]{FFE599}Female Avg} &
  \cellcolor[HTML]{FFE599}\textbf{$_{(83.58\%-)}$16.85\%} &
  \cellcolor[HTML]{FFE599}\hspace{8pt}2.63\% &
  \cellcolor[HTML]{FFE599}\hspace{4pt}92\% &
  \cellcolor[HTML]{FFE599}$_{(5.83)}$-0.23 &
  \cellcolor[HTML]{FFE599}\hspace{4pt}7.72\% &
  \cellcolor[HTML]{FFE599}\hspace{4pt}-3.79\% &
  \cellcolor[HTML]{FFE599}\hspace{4pt}90\% &
  \cellcolor[HTML]{FFE599}-0.11 &
  \cellcolor[HTML]{FFE599}\hspace{4pt}6.42\% &
  \cellcolor[HTML]{FFE599}\hspace{4pt}2.86\% &
  \cellcolor[HTML]{FFE599}\hspace{4pt}90\% &
  \cellcolor[HTML]{FFE599}-0.09 &
  \cellcolor[HTML]{FFE599}\hspace{4pt}7.33\% &
  \cellcolor[HTML]{FFE599}\hspace{4pt}-6.89\% &
  \cellcolor[HTML]{FFE599}\hspace{4pt}92\% &
  \cellcolor[HTML]{FFE599}-0.24 \\ \hline
\rowcolor[HTML]{93C47D} 
\multicolumn{3}{c|}{\cellcolor[HTML]{93C47D}Overall} &
  \textbf{$_{(85.41\%-)}$14.23\%} &
  \hspace{8pt}1.34\% &
  \hspace{4pt}96\% &
  $_{(5.67)}$-0.20 &
  \hspace{4pt}6.53\% &
  \hspace{8pt}7.55\% &
  \hspace{4pt}77\% &
  -0.08 &
  \hspace{4pt}4.73\% &
  \hspace{4pt}7.08\% &
  \hspace{4pt}87\% &
  -0.07 &
  \hspace{4pt}5.04\% &
  \hspace{8pt}4.56\% &
  \hspace{4pt}76\% &
  -0.13 \\ \hline \hline 
\multicolumn{1}{c|}{} &
  \multicolumn{1}{c|}{} &
  hat &
  \textbf{$_{(87.80\%-)}$18.40\%} &
  \hspace{8pt}0.35\% &
  100\% &
  $_{(5.14)}$-0.48 &
  \hspace{4pt}7.90\% &
  \hspace{4pt}16.17\% &
  \hspace{4pt}80\% &
  -0.03 &
  \hspace{4pt}3.60\% &
  \hspace{4pt}1.04\% &
  100\% &
  -0.10 &
  \hspace{4pt}7.80\% &
  \hspace{4pt}15.69\% &
  \hspace{4pt}60\% &
  -0.03 \\
\multicolumn{1}{c|}{} &
  \multicolumn{1}{c|}{} &
  no hat &
  \textbf{$_{(96.00\%-)}$10.00\%} &
  \hspace{4pt}10.11\% &
  \hspace{4pt}80\% &
  $_{(5.43)}$-0.23 &
  \hspace{4pt}6.60\% &
  \hspace{4pt}21.61\% &
  \hspace{4pt}60\% &
  -0.02 &
  \hspace{4pt}2.70\% &
  \hspace{4pt}7.29\% &
  \hspace{4pt}90\% &
  -0.04 &
  \hspace{4pt}6.50\% &
  \hspace{4pt}24.16\% &
  \hspace{4pt}60\% &
  -0.02 \\
\multicolumn{1}{c|}{} &
  \multicolumn{1}{c|}{} &
  glasses &
  \textbf{$_{(95.00\%-)}$24.10\%} &
  \hspace{8pt}0.56\% &
  100\% &
  $_{(5.38)}$-0.12 &
  \hspace{4pt}1.90\% &
  \hspace{8pt}9.16\% &
  \hspace{4pt}80\% &
  -0.10 &
  \hspace{4pt}1.20\% &
  \hspace{4pt}0.39\% &
  100\% &
  -0.01 &
  \hspace{4pt}1.90\% &
  \hspace{8pt}6.48\% &
  \hspace{4pt}90\% &
  -0.10 \\
\multicolumn{1}{c|}{} &
  \multicolumn{1}{c|}{} &
  no glasses &
  \textbf{$_{(94.50\%-)}$23.70\%} &
  \hspace{8pt}2.21\% &
  100\% &
  $_{(5.30)}$+0.02 &
  \hspace{4pt}6.40\% &
  \hspace{4pt}19.10\% &
  \hspace{4pt}70\% &
  -0.00 &
  \hspace{4pt}3.70\% &
  -0.90\% &
  100\% &
  -0.02 &
  \hspace{4pt}6.40\% &
  \hspace{4pt}17.00\% &
  \hspace{4pt}70\% &
  -0.00 \\
\multicolumn{1}{c|}{} &
  \multicolumn{1}{c|}{} &
  oval &
  \textbf{$_{(95.30\%-)}$19.70\%} &
  -10.75\% &
  100\% &
  $_{(5.54)}$-0.16 &
  \hspace{4pt}2.30\% &
  \hspace{4pt}10.22\% &
  100\% &
  -0.12 &
  \hspace{4pt}1.40\% &
  \hspace{4pt}3.20\% &
  100\% &
  -0.11 &
  \hspace{4pt}2.40\% &
  \hspace{8pt}7.82\% &
  100\% &
  -0.12 \\
\multicolumn{1}{c|}{} &
  \multicolumn{1}{c|}{\multirow{-6}{*}{M}} &
  non oval &
  \textbf{$_{(96.40\%-)}$21.30\%} &
  \hspace{8pt}6.86\% &
  100\% &
  $_{(5.32)}$+0.05 &
  \hspace{4pt}6.70\% &
  \hspace{8pt}7.48\% &
  \hspace{4pt}70\% &
  +0.05 &
  \hspace{4pt}5.00\% &
  \hspace{4pt}1.96\% &
  100\% &
  +0.34 &
  \hspace{4pt}6.90\% &
  \hspace{8pt}1.15\% &
  \hspace{4pt}80\% &
  +0.05 \\ \cline{2-19} 
\multicolumn{1}{c|}{} &
  \multicolumn{2}{c|}{\cellcolor[HTML]{FFE599}Male Avg} &
  \cellcolor[HTML]{FFE599}\textbf{$_{(94.17\%-)}$19.53\%} &
  \cellcolor[HTML]{FFE599}\hspace{8pt}1.56\% &
  \cellcolor[HTML]{FFE599}\hspace{4pt}97\% &
  \cellcolor[HTML]{FFE599}$_{(5.35)}$-0.15 &
  \cellcolor[HTML]{FFE599}\hspace{4pt}5.30\% &
  \cellcolor[HTML]{FFE599}\hspace{4pt}13.96\% &
  \cellcolor[HTML]{FFE599}\hspace{4pt}77\% &
  \cellcolor[HTML]{FFE599}-0.04 &
  \cellcolor[HTML]{FFE599}\hspace{4pt}2.93\% &
  \cellcolor[HTML]{FFE599}\hspace{4pt}2.16\% &
  \cellcolor[HTML]{FFE599}\hspace{4pt}98\% &
  \cellcolor[HTML]{FFE599}+0.01 &
  \cellcolor[HTML]{FFE599}\hspace{4pt}5.32\% &
  \cellcolor[HTML]{FFE599}\hspace{4pt}12.05\% &
  \cellcolor[HTML]{FFE599}\hspace{4pt}77\% &
  \cellcolor[HTML]{FFE599}-0.04 \\ \cline{2-19} 
\multicolumn{1}{c|}{} &
  \multicolumn{1}{c|}{} &
  hat &
  \textbf{$_{(93.40\%-)}$21.10\%} &
  \hspace{4pt}-0.14\% &
  \hspace{4pt}80\% &
  $_{(5.33)}$-0.10 &
  -1.10\% &
  -18.35\% &
  100\% &
  -0.10 &
  -1.40\% &
  -4.99\% &
  100\% &
  -0.09 &
  -1.10\% &
  -17.86\% &
  100\% &
  -0.11 \\
\multicolumn{1}{c|}{} &
  \multicolumn{1}{c|}{} &
  no hat &
  \textbf{$_{(92.50\%-)}$17.70\%} &
  \hspace{8pt}2.24\% &
  \hspace{4pt}80\% &
  $_{(6.68)}$-0.44 &
  \hspace{4pt}4.80\% &
  \hspace{4pt}-4.73\% &
  \hspace{4pt}90\% &
  -0.17 &
  \hspace{4pt}4.30\% &
  -5.41\% &
  100\% &
  -0.09 &
  \hspace{4pt}5.00\% &
  \hspace{4pt}-4.82\% &
  \hspace{4pt}90\% &
  -0.17 \\
\multicolumn{1}{c|}{} &
  \multicolumn{1}{c|}{} &
  glasses &
  \textbf{$_{(94.90\%-)}$19.80\%} &
  \hspace{8pt}3.85\% &
  \hspace{4pt}90\% &
  $_{(5.54)}$-0.14 &
  \hspace{4pt}2.90\% &
  \hspace{4pt}-8.75\% &
  100\% &
  +0.01 &
  \hspace{4pt}2.30\% &
  -3.18\% &
  100\% &
  +0.03 &
  \hspace{4pt}3.10\% &
  -10.21\% &
  100\% &
  +0.00 \\
\multicolumn{1}{c|}{} &
  \multicolumn{1}{c|}{} &
  no glasses &
  \textbf{$_{(89.50\%-)}$11.60\%} &
  \hspace{8pt}5.77\% &
  \hspace{4pt}90\% &
  $_{(6.69)}$-0.75 &
  \hspace{4pt}1.10\% &
  \hspace{4pt}-9.04\% &
  \hspace{4pt}90\% &
  -0.32 &
  \hspace{4pt}1.90\% &
  -2.61\% &
  \hspace{4pt}90\% &
  -0.19 &
  \hspace{4pt}1.20\% &
  \hspace{4pt}-9.36\% &
  \hspace{4pt}90\% &
  -0.31 \\
\multicolumn{1}{c|}{} &
  \multicolumn{1}{c|}{} &
  oval &
  \textbf{$_{(84.70\%-)}$24.40\%} &
  \hspace{4pt}-4.28\% &
  100\% &
  $_{(6.00)}$-0.13 &
  \hspace{4pt}4.20\% &
  -16.44\% &
  \hspace{4pt}90\% &
  -0.13 &
  \hspace{4pt}2.60\% &
  -6.86\% &
  100\% &
  -0.13 &
  \hspace{4pt}4.20\% &
  -17.20\% &
  \hspace{4pt}90\% &
  -0.12 \\
\multicolumn{1}{c|}{} &
  \multicolumn{1}{c|}{\multirow{-6}{*}{F}} &
  non oval &
  \textbf{$_{(92.80\%-)}$29.50\%} &
  \hspace{4pt}-1.98\% &
  100\% &
  $_{(5.96)}$-0.19 &
  \hspace{4pt}5.20\% &
  \hspace{4pt}-0.09\% &
  \hspace{4pt}90\% &
  -0.16 &
  \hspace{4pt}3.90\% &
  -2.26\% &
  \hspace{4pt}90\% &
  -0.14 &
  \hspace{4pt}5.10\% &
  \hspace{4pt}-0.39\% &
  \hspace{4pt}90\% &
  -0.16 \\ \cline{2-19} 
\multicolumn{1}{c|}{\multirow{-14}{*}{FFHQ}} &
  \multicolumn{2}{c|}{\cellcolor[HTML]{FFE599}Female Avg} &
  \cellcolor[HTML]{FFE599}\textbf{$_{(91.30\%-)}$20.68\%} &
  \cellcolor[HTML]{FFE599}\hspace{8pt}0.91\% &
  \cellcolor[HTML]{FFE599}\hspace{4pt}90\% &
  \cellcolor[HTML]{FFE599}$_{(6.03)}$-0.26 &
  \cellcolor[HTML]{FFE599}\hspace{4pt}2.85\% &
  \cellcolor[HTML]{FFE599}\hspace{4pt}-9.57\% &
  \cellcolor[HTML]{FFE599}\hspace{4pt}93\% &
  \cellcolor[HTML]{FFE599}-0.15 &
  \cellcolor[HTML]{FFE599}\hspace{4pt}2.27\% &
  \cellcolor[HTML]{FFE599}-4.22\% &
  \cellcolor[HTML]{FFE599}\hspace{4pt}97\% &
  \cellcolor[HTML]{FFE599}-0.10 &
  \cellcolor[HTML]{FFE599}\hspace{4pt}2.92\% &
  \cellcolor[HTML]{FFE599}\hspace{4pt}-9.97\% &
  \cellcolor[HTML]{FFE599}\hspace{4pt}93\% &
  \cellcolor[HTML]{FFE599}-0.14 \\ \hline
\rowcolor[HTML]{93C47D} 
\multicolumn{3}{c|}{\cellcolor[HTML]{93C47D}Overall} &
  \textbf{$_{(92.73\%-)}$20.11\%} &
  \hspace{8pt}1.23\% &
  \hspace{4pt}93\% &
  $_{(5.69)}$-0.27 &
  \hspace{4pt}4.08\% &
  \hspace{8pt}2.20\% &
  \hspace{4pt}85\% &
  -0.09 &
  \hspace{4pt}2.60\% &
  -1.03\% &
  \hspace{4pt}\hspace{4pt}98\% &
  -0.05 &
  \hspace{4pt}4.12\% &
  \hspace{8pt}1.04\% &
  \hspace{4pt}85\% &
  -0.09 \\ \hline
\end{tabular}%
}
\end{table*}

\subsection{Optimization}

We adopt the bi-level optimization algorithm to optimize the architecture parameters $\mathcal{A}$ defined in \secref{subsec:superpert}.
Given a training dataset $\mathcal{D}_\text{train}$ and a validation dataset $\mathcal{D}_\text{valid}$, we have the following functions to optimize
\begin{align}
    \mathcal{B} & = \arg \min_{\mathcal{B}'} \mathcal{L}(\mathcal{B}',\mathcal{A},\mathcal{D}_\text{train},\mathcal{J}),  \label{eq:opt-1} \\ 
    \mathcal{A} & = \arg \min_{\mathcal{A}'} \mathcal{L}(\mathcal{B},\mathcal{A}',\mathcal{D}_\text{val},\mathcal{J}).
    \label{eq:opt-2}
\end{align}
Eq. (\ref{eq:opt-1}) means to optimize the perturbation parameters (\emph{i.e.}, $\mathcal{B}$) with fixed $\mathcal{A}$ to minimize the objective function $\mathcal{J}$ on the training dataset $\mathcal{D}_\text{train}$.
After optimizing $\mathcal{B}$, we fix it and use Eq. (\ref{eq:opt-2}) to optimize the $\mathcal{A}$ by minimizing the objective function $\mathcal{J}$ on the validation dataset $\mathcal{D}_\text{val}$. 
When the $\mathcal{A}$ is optimized, we only preserve one operation for each edge by selecting the operation with maximum weight. 
After fixing the architecture of $f_\beta$, for a given $\mathbf{X}^\text{g}$, we can perform attacks by optimizing the operations' parameters as done in \secref{subsec:naiveimpl}.

\section{Experimental Results}
\subsection{Experimental Setups}
\textbf{Victim Generative Model.} We select the stroke-based image synthesis method SDEdit and utilize a human face model trained on CelebA-HQ dataset as the victim model. Note that our proposed adversarial attack framework is general and can be applied to other stroke-based image synthesis models.

\noindent \textbf{Dataset}. We conduct experiments on two widely-used facial datasets, CelebAMask-HQ \cite{lee2020maskgan} and FFHQ \cite{ffhq}. Inspired by \cite{karkkainenfairface}, we target three sensitive attributes: gender, age, and race with binary value, ``male"/``female", ``young"/``senior", and ``black"/``white" respectively.
To comprehensively evaluate the effectiveness of various attacks and reduce the potential bias stemming from uneven sampling of test data, we further split each attribute into six subgroups according to whether wearing a hat, wearing glasses, and having an oval face (\emph{e.g.}, [female-hat, female-no-hat]). Note that the criteria for subdivision are relatively typical \cite{hand2017attributes}\footnote{Additionally, under the condition of stroke image input, the more intuitive ``whether wearing accessories" and ``oval face" are better choices compared to slight facial features such as ``mouth open" or ``smile".}. We craft the test dataset by selecting 10 samples for each subgroup attribute category.

\begin{figure*}[!hb]
\centering
\begin{minipage}[b]{0.33\textwidth}
\centering
\includegraphics[width=\textwidth]{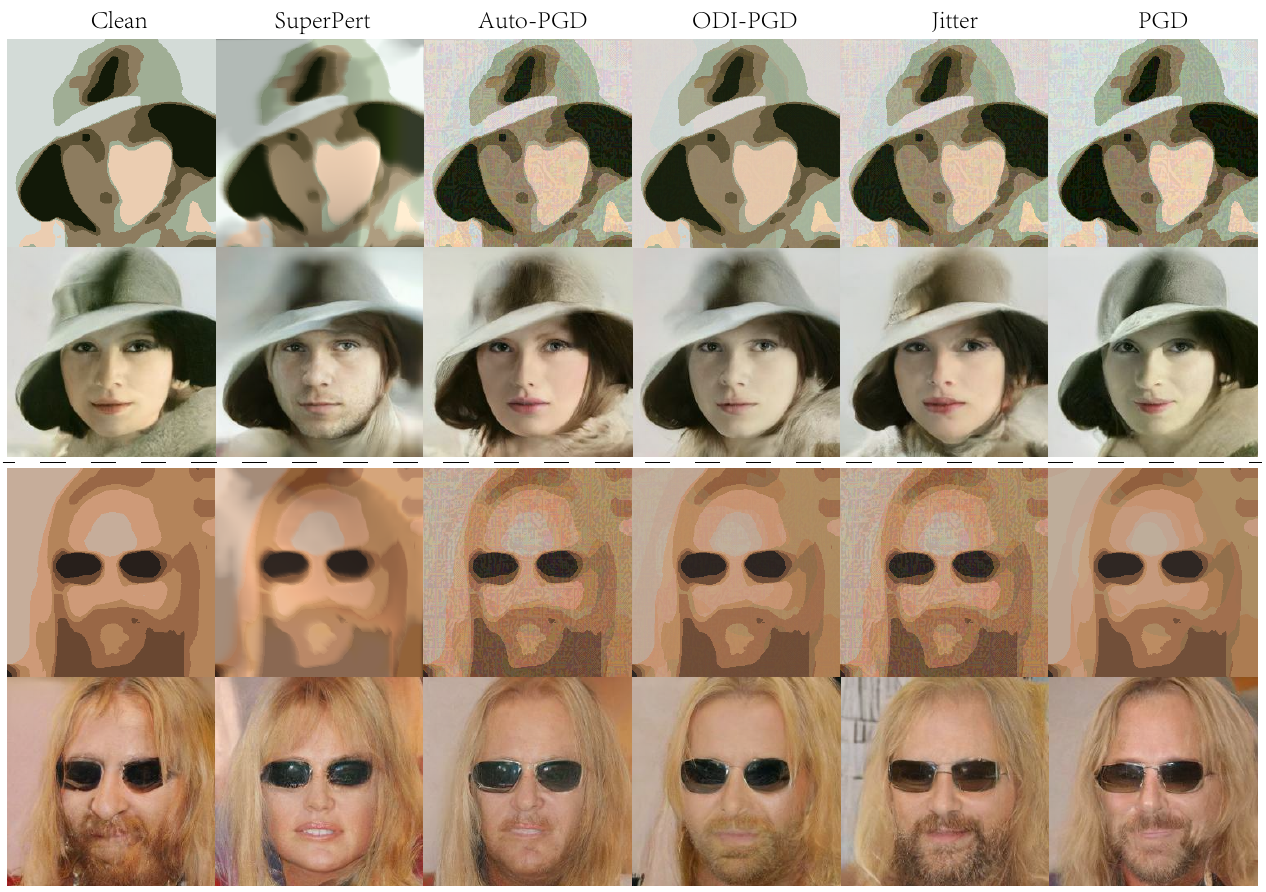}
\subcaption{Top: Female to male transition example, Bottom: Male to female transition example.}
\label{fig:fig_gender}
\end{minipage}
\begin{minipage}[b]{0.33\textwidth}
\centering
\includegraphics[width=\textwidth]{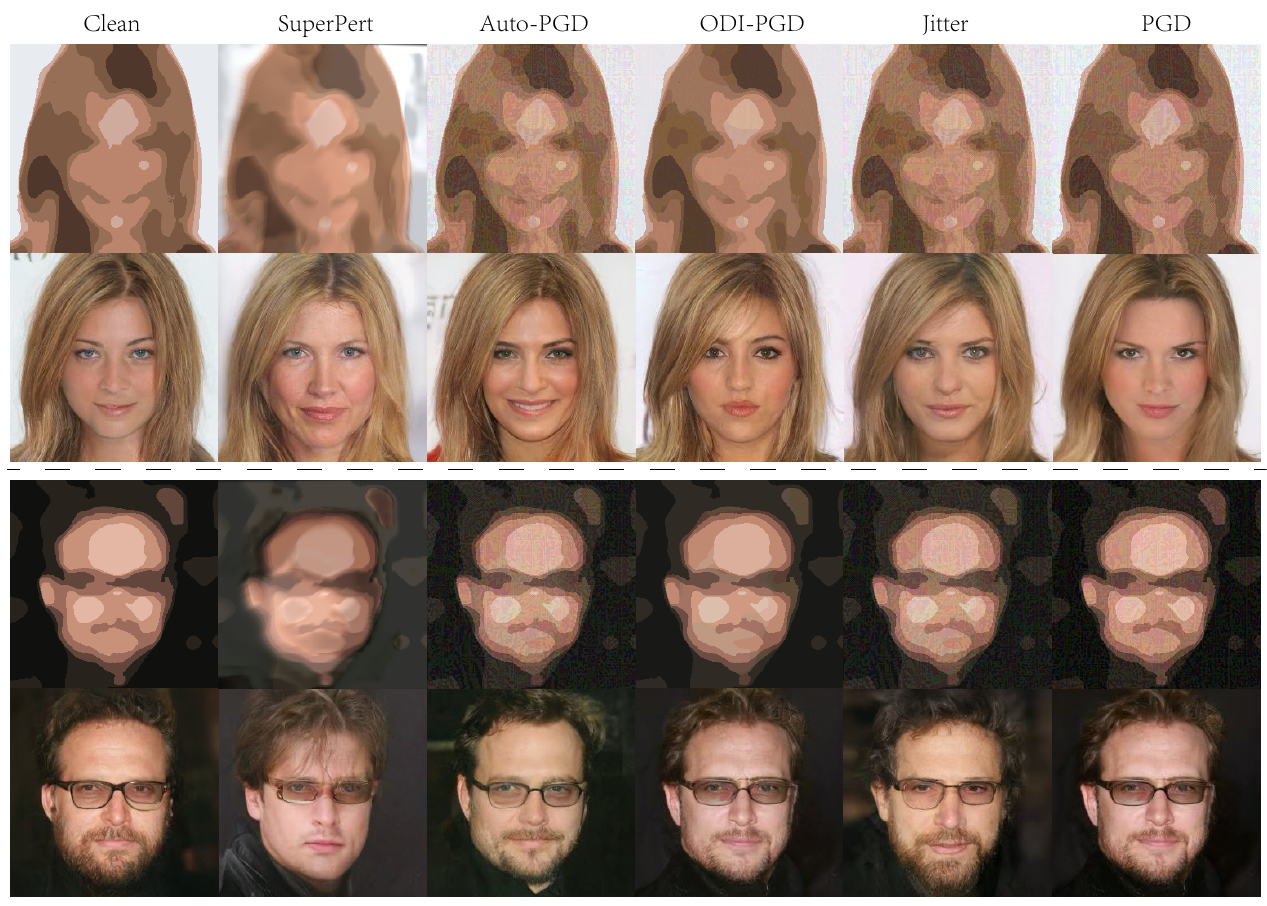}
\subcaption{Top: Young to senior transition example, Bottom: Senior to young transition example.}
\label{fig:fig_age}
\end{minipage}
\begin{minipage}[b]{0.33\textwidth}
\centering
\includegraphics[width=\textwidth]{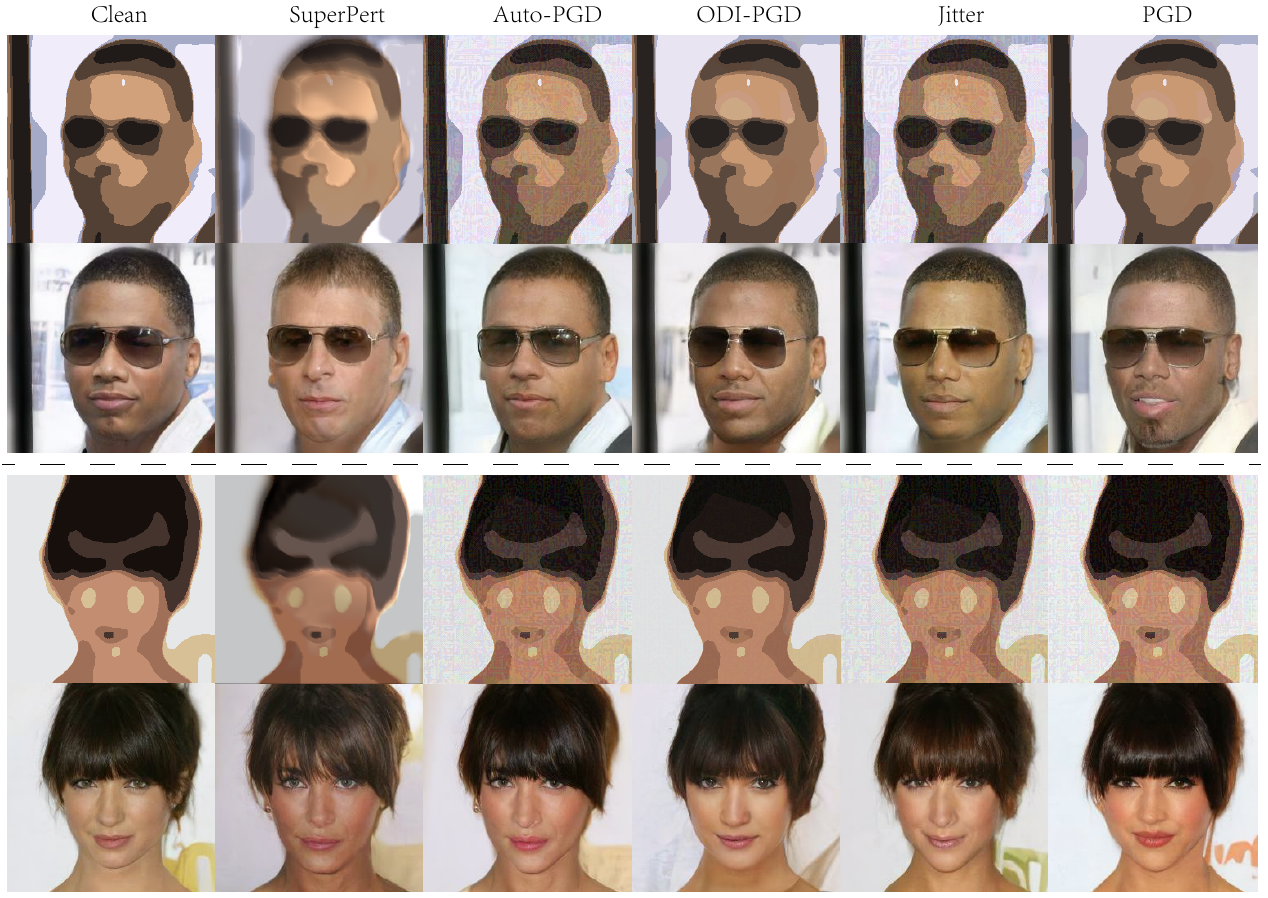}
\subcaption{Top: Black to white transition example, Bottom: White to black transition example.}
\label{fig:fig_race}
\end{minipage}
\vspace{-20pt}
\caption{Visualizations of TAGA: (a), (b), and (c) represent attacks targeting gender attribute, age attribute, and race attribute, respectively. Every first row represents input stroke images and the second row represents generated image samples through victim SDEdit.}
\vspace{-10pt}
\label{fig:merged}
\end{figure*}

\noindent \textbf{Evaluation Metrics}. To reflect the shift of attribute distribution in the generated images, we can compute the average attribute distribution shift as $\text{AADS}(\mathcal{D}_\text{test}) = \frac{1}{N}\sum_{\mathcal{D}_\text{test}}^{\mathbf{X}^{\text{g}}}\frac{\text{TNC}(\tilde{p}_\text{data}) - \text{TNC}({p}_\text{data})}{100}$, where $\text{TNC}(\cdot)$ is the count of the samples with target attribute $\hat{a}$ for each input condition $\mathbf{X}^{\text{g}}$ in the image generation (\emph{i.e.}, $p_\text{data}$) based on $f_{\text{att}}(\cdot,\hat{a})$. For input perturbation evaluation, since our proposed method cannot be effectively measured with the typical norm distance metrics, we instead measure the average input attribute shift targeting clean attribute $a$ after the attack by the average condition attribute distribution shift as $\text{ACADS}(\mathcal{D}_\text{test})=\frac{1}{N}\sum_{\mathcal{D}_\text{test}}^{\mathbf{X}^{\text{g}}}(f_{\text{att}}(\tilde{\mathbf{X}}^{\text{g}}, a)-f_{\text{att}}(\mathbf{X}^{\text{g}}, a))$. And we use the average of condition attribute consistency ratio AoCC to further estimate the input attribute inconsistency rate after the attack. The ideal attack aims to maximize AADS while minimizing ACADS. Therefore, we measured the average attribute change rate ratio as $\text{AACRR}(\mathcal{D}_\text{test})=\frac{1}{N}\sum_{\mathcal{D}_\text{test}}^{\mathbf{X}^{\text{g}}} \frac{\text{TNC}(\tilde{p}_\text{data}) - \text{TNC}({p}_\text{data})}{100\cdot|f_{\text{att}}(\tilde{\mathbf{X}}^{\text{g}}, a)-f_{\text{att}}(\mathbf{X}^{\text{g}}, a)|}$. To demonstrate the changes in generated image quality after the attack, we consider the classic non-reference image quality assessment (IQA) method NIQE \cite{mittal2012making} and the state-of-the-art method TOPIQ \cite{chen2023topiq}\footnote{Noted that the attribute changes induced by TAGA in the attacked generated images compared to the clean generated images, commonly used distribution-based IQA such as IS and FID, along with full-reference IQA methods like PSNR and SSIM, are not appropriate in this scenario.}. Specifically, we compute generated image quality difference as $d_{\text{IQA}}(\mathcal{D}_\text{test}) = \frac{1}{N}\sum_{\mathcal{D}_\text{test}}^{\mathbf{X}^{\text{g}}}\frac{f_\text{IQA}(\tilde{p}_\text{data}) - f_\text{IQA}(p_\text{data})}{100}$. 


%
\noindent \textbf{Baseline Methods}.  
We compare our proposed attack \textit{SuperPert} with four adversarial attacks, PGD \cite{madry2017towards}, ODI-PGD \cite{tashiro2020diversity}, Auto-PGD \cite{croce2020reliable} and Jitter \cite{schwinn2023exploring}. Despite the favorable attack effectiveness of combined approaches (\emph{e.g.}, Auto-Attack \cite{croce2020reliable}) in image classification tasks, the requirement for a substantial number of black-box queries (\emph{e.g.}, $5000$ queries) renders them impractical from a time cost perspective when targeting SDEdit model.


\noindent \textbf{Implementation Details}. \ding{182} For image generation, we follow the default settings of SDEdit with $500$ iterations. For each input, $100$ image samples are generated for evaluation. 
\ding{183} Considering the remarkable zero-shot classification performance reported by CLIP \cite{radford2021learning}, we adopt the CLIP ViT-B/32 for image attribute evaluation. To classify image attributes (\emph{e.g.}, gender, age and race), we use the officially prompt templates "A photo of a \{\} person", with the value sets as \{female, male\}, \{young, senior\}, and \{black, white\}. 
\ding{184} For our method, we train a $10$-layer \textit{SuperPert} on a small dataset of size $20$ selected in CelebAMask-HQ with an ENAS \cite{pham2018efficient} manner to search network architecture. During testing, \textit{SuperPert} performs a round of $20$ iterations PGD optimization targeting given victim input with a learning rate $0.01$. As for baseline methods, we set $L_\infty$ norm restriction $\epsilon=8/255$ and the learning rate $0.01$. In order to maximize the chances of the baseline methods discovering better noise patterns, we set the iteration rounds for all baseline methods to $50$. More detailed settings are in the supplementary materials.
\begin{table*}[!t]
\centering
\caption{The overview results of different attack methods targeting age and race attributes on CelebA dataset.}
\label{tab:table_age_race}
\vspace{-10pt}
\setlength{\tabcolsep}{5.5pt}
\resizebox{1\linewidth}{!}{
\renewcommand\arraystretch{0.99}
\begin{tabular}{ccl|cccc|cccc|cccc|cccc}
\hline
\multicolumn{1}{l}{} &
  \multicolumn{1}{l}{} &
   &
  \multicolumn{4}{c|}{SuperPert} &
  \multicolumn{4}{c|}{Auto-PGD} &
  \multicolumn{4}{c|}{ODI-PGD} &
  \multicolumn{4}{c}{Jitter} \\
\multicolumn{1}{l}{} &
  \multicolumn{1}{l}{} &
   &
  AADS$\textcolor{red}{\uparrow}$ &
  ACADS$\textcolor{blue}{\downarrow}$ &
  AoCC$\textcolor{red}{\uparrow}$ &
  $d_\text{NIQE}\textcolor{blue}{\downarrow}$ &
  AADS$\textcolor{red}{\uparrow}$ &
  ACADS$\textcolor{blue}{\downarrow}$ &
  AoCC$\textcolor{red}{\uparrow}$ &
  $d_\text{NIQE}\textcolor{blue}{\downarrow}$ &
  AADS$\textcolor{red}{\uparrow}$ &
  ACADS$\textcolor{blue}{\downarrow}$ &
  AoCC$\textcolor{red}{\uparrow}$ &
  $d_\text{NIQE}\textcolor{blue}{\downarrow}$ &
  AADS$\textcolor{red}{\uparrow}$ &
  ACADS$\textcolor{blue}{\downarrow}$ &
  AoCC$\textcolor{red}{\uparrow}$ &
  $d_\text{NIQE}\textcolor{blue}{\downarrow}$ \\ \hline \hline 
\multicolumn{1}{c|}{} &
  \multicolumn{1}{c|}{} &
  hat &
  \textbf{$_{(91.70\%-)}$11.10\%} &
  \hspace{4pt}15.65\% &
  \hspace{4pt}70\% &   
  $_{(5.53)}$-0.53 &
  \hspace{4pt}0.20\% &
  \hspace{4pt}-3.34\% &
  \hspace{4pt}90\% &
  -0.17 &
  \hspace{4pt}0.00\% &
  \hspace{4pt}2.93\% &
  100\% &
  -0.15 &
  \hspace{4pt}0.20\% &
  -3.56\% &
  \hspace{4pt}90\% &
  -0.17 \\
\multicolumn{1}{c|}{} &
  \multicolumn{1}{c|}{} &
  no hat &
  \textbf{$_{(75.40\%-)}$\hspace{4pt}7.20\%} &
  -12.18\% &
  \hspace{4pt}90\% &
  $_{(5.83)}$-0.11 &
  \hspace{4pt}3.10\% &
  \hspace{4pt}19.76\% &
  \hspace{4pt}60\% &
  -0.10 &
  \hspace{4pt}0.70\% &
  -0.58\% &
  \hspace{4pt}90\% &
  -0.10 &
  \hspace{4pt}3.20\% &
  19.65\% &
  \hspace{4pt}60\% &
  -0.10 \\
\multicolumn{1}{c|}{} &
  \multicolumn{1}{c|}{} &
  glasses &
  \textbf{$_{(75.20\%-)}$21.00\%} &
  \hspace{8pt}4.65\% &
  100\% &
  $_{(5.33)}$-0.16 &
  \hspace{4pt}9.00\% &
  \hspace{8pt}4.23\% &
  \hspace{4pt}90\% &
  -0.07 &
  \hspace{4pt}9.20\% &
  \hspace{4pt}4.79\% &
  100\% &
  +0.00 &
  10.60\% &
  \hspace{4pt}4.25\% &
  \hspace{4pt}90\% &
  -0.04 \\
\multicolumn{1}{c|}{} &
  \multicolumn{1}{c|}{} &
  no glasses &
  \textbf{$_{(75.70\%-)}$\hspace{4pt}4.00\%} &
  \hspace{4pt}22.88\% &
  \hspace{4pt}80\% &
  $_{(6.11)}$-0.17 &
  \hspace{4pt}1.90\% &
  \hspace{4pt}19.02\% &
  \hspace{4pt}80\% &
  -0.11 &
  \hspace{4pt}0.00\% &
  \hspace{4pt}0.96\% &
  100\% &
  -0.07 &
  \hspace{4pt}3.10\% &
  19.02\% &
  100\% &
  -0.12 \\
\multicolumn{1}{c|}{} &
  \multicolumn{1}{c|}{} &
  oval &
  \textbf{$_{(79.70\%-)}$\hspace{4pt}5.70\%} &
  \hspace{4pt}10.60\% &
  \hspace{4pt}70\% &
  $_{(5.89)}$-0.24 &
  \hspace{4pt}1.10\% &
  \hspace{4pt}11.40\% &
  \hspace{4pt}70\% &
  -0.08 &
  \hspace{4pt}0.10\% &
  -2.70\% &
  100\% &
  -0.08 &
  \hspace{4pt}1.10\% &
  10.84\% &
  \hspace{4pt}70\% &
  -0.08 \\
\multicolumn{1}{c|}{} &
  \multicolumn{1}{c|}{\multirow{-6}{*}{Y}} &
  non oval &
  \textbf{$_{(73.90\%-)}$\hspace{4pt}5.80\%} &
  \hspace{4pt}10.10\% &
  \hspace{4pt}70\% &
  $_{(5.99)}$-0.28 &
  \hspace{4pt}1.20\% &
  \hspace{4pt}11.20\% &
  \hspace{4pt}70\% &
  -0.08 &
  \hspace{4pt}0.30\% &
  -2.73\% &
  100\% &
  -0.08 &
  \hspace{4pt}1.00\% &
  11.10\% &
  \hspace{4pt}70\% &
  -0.08 \\ \cline{2-19} 
\multicolumn{1}{c|}{} &
  \multicolumn{2}{c|}{\cellcolor[HTML]{FFE599}Young Avg} &
  \cellcolor[HTML]{FFE599}\textbf{$_{(78.60\%-)}$\hspace{4pt}9.13\%} &
  \cellcolor[HTML]{FFE599}\hspace{8pt}8.62\% &
  \cellcolor[HTML]{FFE599}\hspace{4pt}80\% &
  \cellcolor[HTML]{FFE599}$_{(5.78)}$-0.25 &
  \cellcolor[HTML]{FFE599}\hspace{4pt}2.75\% &
  \cellcolor[HTML]{FFE599}\hspace{4pt}10.38\% &
  \cellcolor[HTML]{FFE599}\hspace{4pt}77\% &
  \cellcolor[HTML]{FFE599}-0.10 &
  \cellcolor[HTML]{FFE599}\hspace{4pt}1.72\% &
  \cellcolor[HTML]{FFE599}\hspace{4pt}0.45\% &
  \cellcolor[HTML]{FFE599}\hspace{4pt}98\% &
  \cellcolor[HTML]{FFE599}-0.08 &
  \cellcolor[HTML]{FFE599}\hspace{4pt}3.20\% &
  \cellcolor[HTML]{FFE599}10.22\% &
  \cellcolor[HTML]{FFE599}\hspace{4pt}80\% &
  \cellcolor[HTML]{FFE599}-0.10 \\ \cline{2-19} 
\multicolumn{1}{c|}{} &
  \multicolumn{1}{c|}{} &
  hat &
  \textbf{$_{(93.00\%-)}$20.80\%} &
  -20.10\% &
  100\% &
  $_{(5.12)}$+0.03 &
  \hspace{4pt}6.10\% &
  \hspace{8pt}8.56\% &
  \hspace{4pt}70\% &
  -0.06 &
  \hspace{4pt}5.00\% &
  \hspace{4pt}3.64\% &
  100\% &
  -0.05 &
  \hspace{4pt}6.20\% &
  \hspace{4pt}8.89\% &
  \hspace{4pt}70\% &
  -0.06 \\
\multicolumn{1}{c|}{} &
  \multicolumn{1}{c|}{} &
  no hat &
  \textbf{$_{(96.60\%-)}$31.50\%} &
  \hspace{4pt}-6.43\% &
  100\% &
  $_{(5.35)}$+0.25 &
  \hspace{4pt}3.60\% &
  \hspace{4pt}18.63\% &
  \hspace{4pt}80\% &
  +0.10 &
  \hspace{4pt}2.70\% &
  \hspace{4pt}3.97\% &
  100\% &
  +0.09 &
  \hspace{4pt}3.70\% &
  18.87\% &
  \hspace{4pt}70\% &
  +0.10 \\
\multicolumn{1}{c|}{} &
  \multicolumn{1}{c|}{} &
  glasses &
  \textbf{$_{(99.80\%-)}$10.50\%} &
  \hspace{8pt}3.20\% &
  \hspace{4pt}80\% &
  $_{(4.78)}$+0.37 &
  \hspace{4pt}0.30\% &
  \hspace{4pt}10.15\% &
  \hspace{4pt}80\% &
  -0.05 &
  \hspace{4pt}0.00\% &
  \hspace{4pt}4.57\% &
  \hspace{4pt}90\% &
  -0.01 &
  \hspace{4pt}0.00\% &
  \hspace{4pt}9.99\% &
  \hspace{4pt}80\% &
  -0.01 \\
\multicolumn{1}{c|}{} &
  \multicolumn{1}{c|}{} &
  no glasses &
  \textbf{$_{(95.30\%-)}$24.20\%} &
  \hspace{8pt}0.41\% &
  100\% &
  $_{(5.23)}$-0.00 &
  \hspace{4pt}4.90\% &
  \hspace{4pt}17.85\% &
  \hspace{4pt}70\% &
  +0.02 &
  \hspace{4pt}2.90\% &
  \hspace{4pt}4.28\% &
  100\% &
  +0.03 &
  \hspace{4pt}2.60\% &
  18.12\% &
  \hspace{4pt}70\% &
  +0.03 \\
\multicolumn{1}{c|}{} &
  \multicolumn{1}{c|}{} &
  oval &
  \textbf{$_{(74.40\%-)}$38.20\%} &
  \hspace{8pt}2.18\% &
  \hspace{4pt}70\% &
  $_{(5.34)}$+0.22 &
  \hspace{4pt}4.60\% &
  \hspace{4pt}17.54\% &
  \hspace{4pt}40\% &
  +0.20 &
  \hspace{4pt}3.00\% &
  \hspace{4pt}8.25\% &
  \hspace{4pt}60\% &
  +0.17 &
  \hspace{4pt}7.60\% &
  16.82\% &
  \hspace{4pt}50\% &
  +0.19 \\
\multicolumn{1}{c|}{} &
  \multicolumn{1}{c|}{\multirow{-6}{*}{S}} &
  non oval &
  \textbf{$_{(79.00\%-)}$31.10\%} &
  \hspace{8pt}0.24\% &
  \hspace{4pt}90\% &
  $_{(5.26)}$-0.01 &
  \hspace{4pt}8.20\% &
  \hspace{4pt}12.84\% &
  \hspace{4pt}50\% &
  -0.06 &
  \hspace{4pt}5.10\% &
  \hspace{4pt}6.30\% &
  \hspace{4pt}80\% &
  -0.04 &
  \hspace{4pt}7.10\% &
  \hspace{4pt}6.74\% &
  \hspace{4pt}60\% &
  -0.01 \\ \cline{2-19} 
\multicolumn{1}{c|}{\multirow{-14}{*}{\begin{tabular}[c]{@{}c@{}}CelebA\\ Age\end{tabular}}} &
  \multicolumn{2}{c|}{\cellcolor[HTML]{FFE599}Senior Avg} &
  \cellcolor[HTML]{FFE599}\textbf{$_{(89.68\%-)}$26.05\%} &
  \cellcolor[HTML]{FFE599}\hspace{4pt}-3.42\% &
  \cellcolor[HTML]{FFE599}\hspace{4pt}90\% &
  \cellcolor[HTML]{FFE599}$_{(5.18)}$+0.14 &
  \cellcolor[HTML]{FFE599}\hspace{4pt}4.62\% &
  \cellcolor[HTML]{FFE599}\hspace{4pt}14.26\% &
  \cellcolor[HTML]{FFE599}\hspace{4pt}65\% &
  \cellcolor[HTML]{FFE599}+0.02 &
  \cellcolor[HTML]{FFE599}\hspace{4pt}3.12\% &
  \cellcolor[HTML]{FFE599}\hspace{4pt}5.17\% &
  \cellcolor[HTML]{FFE599}\hspace{4pt}88\% &
  \cellcolor[HTML]{FFE599}+0.03 &
  \cellcolor[HTML]{FFE599}\hspace{4pt}4.53\% &
  \cellcolor[HTML]{FFE599}13.24\% &
  \cellcolor[HTML]{FFE599}\hspace{4pt}67\% &
  \cellcolor[HTML]{FFE599}+0.04 \\ \hline
\rowcolor[HTML]{93C47D} 
\multicolumn{3}{c|}{\cellcolor[HTML]{93C47D}Overall} &
  \textbf{$_{(84.63\%-)}$17.59\%} &
  \hspace{8pt}2.60\% &
  \hspace{4pt}85\% &
  $_{(5.47)}$-0.05 &
  \hspace{4pt}3.68\% &
  \hspace{4pt}12.32\% &
  \hspace{4pt}71\% &
  -0.04 &
  \hspace{4pt}2.42\% &
  \hspace{4pt}2.81\% &
  \hspace{4pt}93\% &
  -0.02 &
  \hspace{4pt}3.87\% &
  11.73\% &
  \hspace{4pt}73\% &
  -0.03 \\  \hline \hline 
\multicolumn{1}{c|}{} &
  \multicolumn{1}{c|}{} &
  hat &
  \textbf{$_{(83.20\%-)}$18.80\%} &
  \hspace{4pt}-2.28\% &
  \hspace{4pt}80\% &
  $_{(5.04)}$-0.17 &
  \hspace{4pt}9.20\% &
  \hspace{4pt}15.28\% &
  \hspace{4pt}60\% &
  -0.05 &
  \hspace{4pt}7.00\% &
  \hspace{4pt}1.31\% &
  \hspace{4pt}90\% &
  -0.05 &
  \hspace{4pt}9.20\% &
  15.01\% &
  \hspace{4pt}60\% &
  -0.06 \\
\multicolumn{1}{c|}{} &
  \multicolumn{1}{c|}{} &
  no hat &
  \textbf{$_{(67.20\%-)}$26.60\%} &
  \hspace{4pt}-7.70\% &
  \hspace{4pt}80\% &
  $_{(5.60)}$-0.39 &
  10.80\% &
  \hspace{4pt}13.62\% &
  \hspace{4pt}30\% &
  -0.06 &
  \hspace{4pt}7.20\% &
  \hspace{4pt}5.16\% &
  \hspace{4pt}70\% &
  -0.04 &
  10.30\% &
  12.99\% &
  \hspace{4pt}30\% &
  -0.06 \\
\multicolumn{1}{c|}{} &
  \multicolumn{1}{c|}{} &
  glasses &
  \textbf{$_{(76.80\%-)}$31.40\%} &
  \hspace{4pt}-0.33\% &
  \hspace{4pt}70\% &
  $_{(5.09)}$+0.01 &
  12.60\% &
  \hspace{8pt}5.66\% &
  \hspace{4pt}50\% &
  -0.06 &
  11.10\% &
  -1.23\% &
  \hspace{4pt}70\% &
  -0.05 &
  13.00\% &
  \hspace{4pt}6.01\% &
  \hspace{4pt}50\% &
  -0.06 \\
\multicolumn{1}{c|}{} &
  \multicolumn{1}{c|}{} &
  no glasses &
  \textbf{$_{(73.50\%-)}$30.80\%} &
  \hspace{4pt}-5.46\% &
  \hspace{4pt}80\% &
  $_{(5.99)}$-0.38 &
  \hspace{4pt}5.30\% &
  \hspace{8pt}9.15\% &
  \hspace{4pt}60\% &
  -0.07 &
  \hspace{4pt}2.40\% &
  \hspace{4pt}1.81\% &
  \hspace{4pt}80\% &
  -0.08 &
  \hspace{4pt}5.10\% &
  \hspace{4pt}8.51\% &
  \hspace{4pt}50\% &
  -0.07 \\
\multicolumn{1}{c|}{} &
  \multicolumn{1}{c|}{} &
  oval &
  \textbf{$_{(55.30\%-)}$29.50\%} &
  \hspace{4pt}-9.19\% &
  \hspace{4pt}90\% &
  $_{(6.24)}$-0.39 &
  10.10\% &
  \hspace{8pt}2.95\% &
  \hspace{4pt}40\% &
  -0.18 &
  10.40\% &
  -2.21\% &
  100\% &
  -0.16 &
  10.30\% &
  \hspace{4pt}2.33\% &
  \hspace{4pt}40\% &
  -0.18 \\
\multicolumn{1}{c|}{} &
  \multicolumn{1}{c|}{\multirow{-6}{*}{B}} &
  non oval &
  \textbf{$_{(69.10\%-)}$27.30\%} &
  \hspace{4pt}-8.72\% &
  100\% &
  $_{(5.83)}$-0.45 &
  \hspace{4pt}4.30\% &
  \hspace{8pt}6.29\% &
  \hspace{4pt}70\% &
  -0.24 &
  \hspace{4pt}2.00\% &
  \hspace{4pt}0.46\% &
  \hspace{4pt}90\% &
  -0.22 &
  \hspace{4pt}4.00\% &
  \hspace{4pt}6.12\% &
  \hspace{4pt}60\% &
  -0.24 \\ \cline{2-19} 
\multicolumn{1}{c|}{} &
  \multicolumn{2}{c|}{\cellcolor[HTML]{FFE599}Black Avg} &
  \cellcolor[HTML]{FFE599}\textbf{$_{(70.85\%-)}$27.40\%} &
  \cellcolor[HTML]{FFE599}\hspace{4pt}-5.61\% &
  \cellcolor[HTML]{FFE599}\hspace{4pt}83\% &
  \cellcolor[HTML]{FFE599}$_{(5.63)}$-0.30 &
  \cellcolor[HTML]{FFE599}\hspace{4pt}8.72\% &
  \cellcolor[HTML]{FFE599}\hspace{8pt}8.83\% &
  \cellcolor[HTML]{FFE599}\hspace{4pt}52\% &
  \cellcolor[HTML]{FFE599}-0.11 &
  \cellcolor[HTML]{FFE599}\hspace{4pt}6.68\% &
  \cellcolor[HTML]{FFE599}\hspace{4pt}0.88\% &
  \cellcolor[HTML]{FFE599}\hspace{4pt}83\% &
  \cellcolor[HTML]{FFE599}-0.10 &
  \cellcolor[HTML]{FFE599}\hspace{4pt}8.65\% &
  \cellcolor[HTML]{FFE599}\hspace{4pt}8.50\% &
  \cellcolor[HTML]{FFE599}\hspace{4pt}48\% &
  \cellcolor[HTML]{FFE599}-0.11 \\ \cline{2-19} 
\multicolumn{1}{c|}{} &
  \multicolumn{1}{c|}{} &
  hat &
  \textbf{$_{(97.10\%-)}$\hspace{4pt}2.70\%} &
  \hspace{8pt}3.90\% &
  \hspace{4pt}80\% &
  $_{(6.13)}$-0.42 &
  \hspace{4pt}0.30\% &
  \hspace{8pt}6.16\% &
  \hspace{4pt}80\% &
  -0.16 &
  \hspace{4pt}0.30\% &
  \hspace{4pt}0.40\% &
  \hspace{4pt}80\% &
  -0.16 &
  \hspace{4pt}0.30\% &
  \hspace{4pt}6.15\% &
  \hspace{4pt}70\% &
  -0.16 \\
\multicolumn{1}{c|}{} &
  \multicolumn{1}{c|}{} &
  no hat &
  \textbf{$_{(96.80\%-)}$\hspace{4pt}5.10\%} &
  \hspace{4pt}-3.49\% &
  \hspace{4pt}80\% &
  $_{(6.11)}$-0.25 &
  \hspace{4pt}2.60\% &
  \hspace{8pt}2.92\% &
  \hspace{4pt}80\% &
  +0.02 &
  \hspace{4pt}1.60\% &
  -2.45\% &
  100\% &
  +0.04 &
  \hspace{4pt}2.60\% &
  -3.15\% &
  \hspace{4pt}80\% &
  +0.02 \\
\multicolumn{1}{c|}{} &
  \multicolumn{1}{c|}{} &
  glasses &
  \textbf{$_{(94.60\%-)}$13.00\%} &
  \hspace{8pt}1.33\% &
  \hspace{4pt}90\% &
  $_{(4.95)}$+0.16 &
  \hspace{4pt}4.60\% &
  \hspace{8pt}7.28\% &
  \hspace{4pt}60\% &
  +0.00 &
  \hspace{4pt}3.40\% &
  \hspace{4pt}3.05\% &
  \hspace{4pt}60\% &
  +0.01 &
  \hspace{4pt}4.50\% &
  \hspace{4pt}7.15\% &
  \hspace{4pt}60\% &
  +0.00 \\
\multicolumn{1}{c|}{} &
  \multicolumn{1}{c|}{} &
  no glasses &
  \textbf{$_{(97.20\%-)}$\hspace{4pt}3.50\%} &
  \hspace{8pt}3.31\% &
  \hspace{4pt}70\% &
  $_{(5.39)}$+0.26 &
  2.40\% &
  \hspace{8pt}6.57\% &
  \hspace{4pt}40\% &
  -0.01 &
  \hspace{4pt}1.60\% &
  -2.35\% &
  \hspace{4pt}70\% &
  +0.01 &
  \hspace{4pt}2.40\% &
  \hspace{4pt}6.59\% &
  \hspace{4pt}40\% &
  +0.01 \\
\multicolumn{1}{c|}{} &
  \multicolumn{1}{c|}{} &
  oval &
  \textbf{$_{(75.90\%-)}$\hspace{4pt}5.00\%} &
  \hspace{4pt}-2.75\% &
  \hspace{4pt}90\% &
  $_{(6.92)}$-0.53 &
  \hspace{4pt}1.20\% &
  \hspace{8pt}4.97\% &
  \hspace{4pt}60\% &
  -0.09 &
  \hspace{4pt}3.10\% &
  -0.50\% &
  \hspace{4pt}90\% &
  -0.09 &
  \hspace{4pt}1.30\% &
  \hspace{4pt}4.68\% &
  \hspace{4pt}50\% &
  -0.09 \\
\multicolumn{1}{c|}{} &
  \multicolumn{1}{c|}{\multirow{-6}{*}{W}} &
  non oval &
  \textbf{$_{(97.40\%-)}$\hspace{4pt}4.30\%} &
  \hspace{8pt}0.88\% &
  \hspace{4pt}90\% &
  $_{(5.91)}$-0.30 &
  \hspace{4pt}0.90\% &
  -13.70\% &
  \hspace{4pt}80\% &
  -0.05 &
  \hspace{4pt}0.60\% &
  -0.13\% &
  \hspace{4pt}70\% &
  -0.04 &
  \hspace{4pt}1.10\% &
  \hspace{4pt}2.22\% &
  \hspace{4pt}50\% &
  -0.05 \\ \cline{2-19} 
\multicolumn{1}{c|}{\multirow{-14}{*}{\begin{tabular}[c]{@{}c@{}}CelebA\\ Race\end{tabular}}} &
  \multicolumn{2}{c|}{\cellcolor[HTML]{FFE599}White Avg} &
  \cellcolor[HTML]{FFE599}\textbf{$_{(93.17\%-)}$\hspace{4pt}5.60\%} &
  \cellcolor[HTML]{FFE599}\hspace{8pt}0.53\% &
  \cellcolor[HTML]{FFE599}\hspace{4pt}83\% &
  \cellcolor[HTML]{FFE599}$_{(5.90)}$-0.18 &
  \cellcolor[HTML]{FFE599}\hspace{4pt}2.00\% &
  \cellcolor[HTML]{FFE599}\hspace{8pt}2.37\% &
  \cellcolor[HTML]{FFE599}\hspace{4pt}67\% &
  \cellcolor[HTML]{FFE599}-0.05 &
  \cellcolor[HTML]{FFE599}\hspace{4pt}1.77\% &
  \cellcolor[HTML]{FFE599}-0.33\% &
  \cellcolor[HTML]{FFE599}\hspace{4pt}78\% &
  \cellcolor[HTML]{FFE599}-0.04 &
  \cellcolor[HTML]{FFE599}\hspace{4pt}2.03\% &
  \cellcolor[HTML]{FFE599}\hspace{4pt}3.94\% &
  \cellcolor[HTML]{FFE599}\hspace{4pt}58\% &
  \cellcolor[HTML]{FFE599}-0.04 \\ \hline
\rowcolor[HTML]{93C47D} 
\multicolumn{3}{c|}{\cellcolor[HTML]{93C47D}Overall} &
  \textbf{$_{(82.01\%-)}$16.50\%} &
  \hspace{4pt}-2.54\% &
  \hspace{4pt}83\% &
  $_{(5.77)}$-0.24 &
  \hspace{4pt}5.36\% &
  \hspace{8pt}5.60\% &
  \hspace{4pt}59\% &
  -0.08 &
  \hspace{4pt}4.23\% &
  \hspace{4pt}0.28\% &
  \hspace{4pt}81\% &
  -0.07 &
  \hspace{4pt}5.34\% &
  \hspace{4pt}6.22\% &
  \hspace{4pt}53\% &
  -0.08 \\ \hline
\end{tabular}%
}
\end{table*}

\begin{table*}[!t]
\centering
\caption{The result of ablation study targeting gender attribute on CelebA dataset.}
\label{tab:table_ablation}
\vspace{-10pt}

\setlength{\tabcolsep}{5.5pt}
\resizebox{1\linewidth}{!}{
\renewcommand\arraystretch{0.97}
\begin{tabular}{ccl|cccc|cccc|cccc|cccc}
\hline
\multicolumn{1}{l}{}                           & \multicolumn{1}{l}{}                      &             & \multicolumn{4}{c|}{Single-layer Exposure}                                                                                     & \multicolumn{4}{c|}{Single-layer Blur}                                                                                           & \multicolumn{4}{c|}{Multi-layer Exposure}                                                                                       & \multicolumn{4}{c}{Multi-layer Blur}                                                                                             \\
\multicolumn{1}{l}{}                           & \multicolumn{1}{l}{}                      &             & AADS$\textcolor{red}{\uparrow}$                           & ACADS$\textcolor{blue}{\downarrow}$                          & AoCC$\textcolor{red}{\uparrow}$                         & $d_\text{NIQE}\textcolor{blue}{\downarrow}$               & AADS$\textcolor{red}{\uparrow}$                           & ACADS$\textcolor{blue}{\downarrow}$                           & AoCC$\textcolor{red}{\uparrow}$                          & $d_\text{NIQE}\textcolor{blue}{\downarrow}$               & AADS$\textcolor{red}{\uparrow}$                           & ACADS$\textcolor{blue}{\downarrow}$                          & AoCC$\textcolor{red}{\uparrow}$                          & $d_\text{NIQE}\textcolor{blue}{\downarrow}$               & AADS$\textcolor{red}{\uparrow}$                            & ACADS$\textcolor{blue}{\downarrow}$                           & AoCC$\textcolor{red}{\uparrow}$                         & $d_\text{NIQE}\textcolor{blue}{\downarrow}$               \\ \hline \hline
\multicolumn{1}{c|}{}                          & \multicolumn{1}{c|}{}                     & hat         & \hspace{4pt}0.50\%                         & \hspace{4pt}0.28\%                         & 100\%                        & +0.28                         & 1.00\%                         & -5.11\%                         & 100\%                         & +0.14                         & \hspace{4pt}1.90\%                         & -0.84\%                        & 100\%                         & +0.39                         & \hspace{4pt}4.80\%                          & \hspace{4pt}3.72\%                          & 100\%                        & +0.06                         \\
\multicolumn{1}{c|}{}                          & \multicolumn{1}{c|}{}                     & no hat      & \hspace{4pt}4.00\%                         & \hspace{4pt}2.91\%                         & 100\%                        & +0.07                         & 1.50\%                         & -0.58\%                         & 100\%                         & +0.08                         & \hspace{4pt}7.50\%                         & \hspace{4pt}2.34\%                         & 100\%                         & +0.10                         & \hspace{4pt}3.30\%                          & -5.23\%                         & \hspace{4pt}90\%                         & +0.17                         \\
\multicolumn{1}{c|}{}                          & \multicolumn{1}{c|}{}                     & glasses     & \hspace{4pt}5.60\%                         & -0.27\%                        & 100\%                        & +0.05                         & 2.10\%                         & -3.40\%                         & 100\%                         & +0.01                         & \hspace{4pt}7.60\%                         & -0.50\%                        & 100\%                         & +0.11                         & \hspace{4pt}2.90\%                          & \hspace{4pt}0.31\%                          & 100\%                        & +0.05                         \\
\multicolumn{1}{c|}{}                          & \multicolumn{1}{c|}{}                     & no glasses  & \hspace{4pt}2.50\%                         & -0.34\%                        & 100\%                        & +0.05                         & 3.50\%                         & \hspace{4pt}1.42\%                          & 100\%                         & +0.03                         & \hspace{4pt}5.10\%                         & \hspace{4pt}1.88\%                         & 100\%                         & +0.08                         & \hspace{4pt}3.70\%                          & \hspace{4pt}1.62\%                          & 100\%                        & +0.08                         \\
\multicolumn{1}{c|}{}                          & \multicolumn{1}{c|}{}                     & oval        & \hspace{4pt}4.50\%                         & \hspace{4pt}0.22\%                         & 100\%                        & +0.00                         & 2.30\%                         & -3.75\%                         & 100\%                         & +0.01                         & \hspace{4pt}9.20\%                         & \hspace{4pt}0.18\%                         & 100\%                         & +0.03                         & \hspace{4pt}1.70\%                          & -7.70\%                         & 100\%                        & +0.09                         \\
\multicolumn{1}{c|}{}                          & \multicolumn{1}{c|}{\multirow{-6}{*}{M}}  & non oval    & \hspace{4pt}2.50\%                         & \hspace{4pt}2.72\%                         & \hspace{4pt}90\%                         & +0.03                         & 0.40\%                         & \hspace{4pt}0.73\%                          & 100\%                         & +0.06                         & \hspace{4pt}6.00\%                         & \hspace{4pt}2.52\%                         & 100\%                         & +0.05                         & \hspace{4pt}2.20\%                          & \hspace{4pt}6.76\%                          & \hspace{4pt}90\%                         & +0.07                         \\ \cline{2-19} 
\multicolumn{1}{c|}{}                          & \multicolumn{2}{c|}{\cellcolor[HTML]{FFE599}Male Avg}   & \cellcolor[HTML]{FFE599}\hspace{4pt}3.27\% & \cellcolor[HTML]{FFE599}\hspace{4pt}0.92\% & \cellcolor[HTML]{FFE599}\hspace{4pt}98\% & \cellcolor[HTML]{FFE599}+0.08 & \cellcolor[HTML]{FFE599}1.80\% & \cellcolor[HTML]{FFE599}-1.78\% & \cellcolor[HTML]{FFE599}100\% & \cellcolor[HTML]{FFE599}+0.06 & \cellcolor[HTML]{FFE599}\hspace{4pt}6.22\% & \cellcolor[HTML]{FFE599}\hspace{4pt}0.93\% & \cellcolor[HTML]{FFE599}100\% & \cellcolor[HTML]{FFE599}+0.13 & \cellcolor[HTML]{FFE599}\hspace{4pt}3.10\%  & \cellcolor[HTML]{FFE599}-0.09\% & \cellcolor[HTML]{FFE599}\hspace{4pt}97\% & \cellcolor[HTML]{FFE599}+0.09 \\ \cline{2-19} 
\multicolumn{1}{c|}{}                          & \multicolumn{1}{c|}{}                     & hat         & \hspace{4pt}1.50\%                         & \hspace{4pt}3.90\%                         & \hspace{4pt}90\%                         & +0.05                         & 0.20\%                         & \hspace{4pt}2.16\%                          & 100\%                         & +0.04                         & \hspace{4pt}7.70\%                         & \hspace{4pt}4.47\%                         & 100\%                         & +0.12                         & 12.60\%                         & -5.55\%                         & 100\%                        & +0.24                         \\
\multicolumn{1}{c|}{}                          & \multicolumn{1}{c|}{}                     & no hat      & \hspace{4pt}6.00\%                         & \hspace{4pt}0.39\%                         & \hspace{4pt}90\%                         & +0.12                         & 4.80\%                         & \hspace{4pt}0.18\%                          & \hspace{4pt}90\%                          & +0.12                         & 12.00\%                        & \hspace{4pt}0.50\%                         & 100\%                         & +0.16                         & 13.00\%                         & \hspace{4pt}9.28\%                          & \hspace{4pt}90\%                         & +0.33                         \\
\multicolumn{1}{c|}{}                          & \multicolumn{1}{c|}{}                     & glasses     & \hspace{4pt}7.10\%                         & \hspace{4pt}2.73\%                         & \hspace{4pt}90\%                         & +0.04                         & 4.90\%                         & \hspace{4pt}2.91\%                          & 100\%                         & +0.02                         & 10.10\%                        & \hspace{4pt}3.13\%                         & 100\%                         & +0.07                       & 19.10\%                         & \hspace{4pt}7.40\%                          & \hspace{4pt}80\%                         & +0.38                         \\
\multicolumn{1}{c|}{}                          & \multicolumn{1}{c|}{}                     & no glasses  & 11.30\%                        & \hspace{4pt}5.51\%                         & \hspace{4pt}80\%                         & +0.10                         & 8.90\%                         & -7.26\%                         & \hspace{4pt}90\%                          & +0.08                         & 11.40\%                        & \hspace{4pt}3.62\%                         & \hspace{4pt}80\%                          & +0.08                         & 15.40\%                         & -2.54\%                         & \hspace{4pt}80\%                         & +0.13                         \\
\multicolumn{1}{c|}{}                          & \multicolumn{1}{c|}{}                     & oval        & \hspace{4pt}0.80\%                         & -0.79\%                        & \hspace{4pt}90\%                         & -0.09                        & 0.10\%                         & -5.76\%                         & 100\%                         & +0.45                         & \hspace{4pt}1.70\%                         & \hspace{4pt}0.15\%                         & 100\%                         & -0.11                        & \hspace{4pt}4.20\%                          & -1.98\%                         & 100\%                        & +0.61                         \\
\multicolumn{1}{c|}{}                          & \multicolumn{1}{c|}{\multirow{-6}{*}{F}}  & non oval    & \hspace{4pt}5.70\%                         & \hspace{4pt}6.03\%                         & \hspace{4pt}90\%                         & +0.12                         & 5.30\%                         & -8.03\%                         & \hspace{4pt}90\%                          & +0.16                         & \hspace{4pt}7.60\%                         & \hspace{4pt}5.44\%                         & \hspace{4pt}80\%                          & +0.16                         & \hspace{4pt}6.00\%                          & -0.69\%                         & \hspace{4pt}80\%                         & +0.34                         \\ \cline{2-19} 
\multicolumn{1}{c|}{\multirow{-14}{*}{CelebA}} & \multicolumn{2}{c|}{\cellcolor[HTML]{FFE599}Female Avg} & \cellcolor[HTML]{FFE599}\hspace{4pt}5.40\% & \cellcolor[HTML]{FFE599}\hspace{4pt}2.96\% & \cellcolor[HTML]{FFE599}\hspace{4pt}88\% & \cellcolor[HTML]{FFE599}+0.06 & \cellcolor[HTML]{FFE599}4.03\% & \cellcolor[HTML]{FFE599}-2.63\% & \cellcolor[HTML]{FFE599}\hspace{4pt}95\%  & \cellcolor[HTML]{FFE599}+0.15 & \cellcolor[HTML]{FFE599}\hspace{4pt}8.42\% & \cellcolor[HTML]{FFE599}\hspace{4pt}2.89\% & \cellcolor[HTML]{FFE599}\hspace{4pt}93\%  & \cellcolor[HTML]{FFE599}+0.08 & \cellcolor[HTML]{FFE599}11.72\% & \cellcolor[HTML]{FFE599}\hspace{4pt}0.99\%  & \cellcolor[HTML]{FFE599}\hspace{4pt}88\% & \cellcolor[HTML]{FFE599}+0.34 \\ \hline
\rowcolor[HTML]{93C47D} 
\multicolumn{3}{c|}{\cellcolor[HTML]{93C47D}Overall}                                                     & \hspace{4pt}4.33\%                         & \hspace{4pt}1.94\%                         & \hspace{4pt}93\%                         & +0.07                         & 2.92\%                         & -2.21\%                         & \hspace{4pt}98\%                          & +0.10                         & \hspace{4pt}7.32\%                         & \hspace{4pt}1.91\%                         & \hspace{4pt}97\%                          & +0.10                         & \hspace{4pt}7.41\%                          & \hspace{4pt}0.45\%                          & \hspace{4pt}93\%                         & +0.21                         \\ \hline
\end{tabular}
}
\end{table*}
\subsection{Attack on Gender Attribute}
We first conduct experiments targeting the gender attribute using the CelebA and FFHQ datasets with $12$ subgroups.

In all groups of two datasets, our proposed method demonstrates superior attack effectiveness compared to the four baseline methods, as shown in \tableref{tab:table_gender}. For instance, our method achieves an average attribute distribution shift of $14.23\%$ on CelebA and $20.11\%$ on FFHQ. In the baseline approach, Auto-PGD and Jitter exhibits relatively better adversarial effects. However, both methods introduce more modifications to the original condition attributes, and their ability to maintain AoCC is slightly inferior to other methods. As reflected in the ACADS, our method demonstrates the distribution shift of $1.34\%$ on CelebA dataset, while the Auto-PGD and Jitter exhibit moderate attack effects but with modification magnitudes of $7.55\%$ and $4.56\%$ respectively.
Additionally, in terms of image quality, all experimental methods are comparable to the quality of the original generated images. A visualization result of each attack method is shown in \figref{fig:fig_gender}. More metrics to show the attribute change rate ratio and image quality are in the supplementary materials.

\subsection{Attack on Age and Race Attributes}
We further evaluate the effectiveness of various TAGA methods targeting age and race attributes. As the results are shown in \tableref{tab:table_age_race}, our method also demonstrates superior attack performance across all groups. It can be observed that the attack effectiveness against the ``young" and ``white" groups is generally weaker compared to the ``senior" and ``black" groups. For instance, when evaluated on the CelebA dataset, SuperPert demonstrates AADS of $9.13\%$ and $26.05\%$ for the ``young" and ``senior" groups, $5.60\%$ and $27.40\%$ for ``white" and ``black" attribute on CelebA dataset, while similar results are also demonstrated on the FFHQ dataset in supplementary materials. We hypothesize that the SDEdit model trained on the CelebA dataset has a bias towards white skin tones and young groups. The illustration samples are shown in \figref{fig:fig_age} and \figref{fig:fig_race}.

\subsection{Ablation Study}
To assess the impact of each adversarial module on the effectiveness of the attack, we divided the adversarial exposure and blurring modules into separate components to create four models: a single-layer exposure model, a single-layer blur model, a multi-layer exposure model, and a multi-layer blur model.  Experimental Result is shown in \tableref{tab:table_ablation}. While single-layer exposure and blur have noticeable effects in certain groups, the overall performance of the multi-layer exposure and multi-layer blur models is more pronounced. Both of them make notable contributions to the adversarial effect.





\section{Conclusion}
In this paper, we introduce a novel attack targeting the guided image synthesis method SDEdit, to prompt it to generate a distribution aligned with specific attributes. This task is formulated as the Targeted Attribute Generative Attack (TAGA).
%
Through empirical studies, we observed that traditional adversarial additive noise attacks fall short of achieving the attack objective. Additionally, we discovered that natural perturbations, such as exposure effects and motion blur, can easily influence the attributes of the generated images.
Motivated by these observations, we propose \textsc{FoolSDEdit}, which incorporates an optimized execution strategy for various natural perturbations, ensuring the effective execution of TAGA against SDEdit.
We validate the effectiveness of our method on two datasets and three attributes, unveiling the potential vulnerability of SDEdit. Given the widespread usage of SDEdit across different tasks, addressing this vulnerability is crucial and may drive subsequent research.

\bibliographystyle{named}
\bibliography{ijcai24}

\end{document}